\documentclass[conference]{IEEEtran}
\IEEEoverridecommandlockouts
\usepackage{cite}
\usepackage{amsmath,amssymb,amsfonts}
\usepackage{algorithmic}
\usepackage{graphicx}
\usepackage{textcomp}
\usepackage{xcolor}
\usepackage{booktabs}
\usepackage{color}
\usepackage{algorithm,algorithmic}
\usepackage{subcaption}
\newcommand{\ie}{{i.e.}}
\newcommand{\eg}{{e.g.}}
\usepackage{array}

\usepackage{lipsum,graphicx,multicol}

\newcolumntype{H}{>{\setbox0=\hbox\bgroup}c<{\egroup}@{}}

\def\BibTeX{{\rm B\kern-.05em{\sc i\kern-.025em b}\kern-.08em
    T\kern-.1667em\lower.7ex\hbox{E}\kern-.125emX}}
\begin{document}

\title{Friendly Training: Neural Networks Can \\ Adapt Data To Make Learning Easier
\thanks{This work was partly supported by the PRIN 2017 project RexLearn, funded by the Italian Ministry of Education, University and Research (grant no. 2017TWNMH2).

Accepted  for  publication  at  the  IEEE  International  Joint  Conference  on Neural Networks (IJCNN) 2021 (DOI: TBA).

2021  IEEE.  Personal  use  of  this  material  is  permitted.  Permission  from IEEE  must  be  obtained  for  all  other  uses,  in  any  current  or  future  media, including reprinting/republishing this material for advertising or promotional purposes, creating new collective works, for resale or redistribution to servers or lists, or reuse of any copyrighted component of this work in other works.}
}

\author{

\IEEEauthorblockN{\hskip 3.5cm Simone Marullo\IEEEauthorrefmark{1}\IEEEauthorrefmark{2},}
\IEEEauthorblockA{\IEEEauthorrefmark{1}\textit{Dept. of Information Engineering} \\
\textit{University of Florence}\\
Florence, Italy}
\and
\IEEEauthorblockN{\hskip -3cm Matteo Tiezzi\IEEEauthorrefmark{2}, Marco Gori\IEEEauthorrefmark{2}\IEEEauthorrefmark{3},}
\IEEEauthorblockA{\IEEEauthorrefmark{2}\textit{Dept. of Information Engineering and Mathematics} \\
\textit{University of Siena}\\
Siena, Italy \\
 \hskip -3cm \small{\texttt{simone.marullo@unifi.it}},
 \small{\texttt{\{mtiezzi,marco,mela\}@diism.unisi.it}}}
\and
\IEEEauthorblockN{\hskip -6.6cm Stefano Melacci\IEEEauthorrefmark{2}}
\IEEEauthorblockA{\IEEEauthorrefmark{3}\textit{MAASAI} \\
\textit{Universit\`{e} C\^{o}te d'Azur,}\\
Nice, France}
}

\maketitle

\begin{abstract}
In the last decade, motivated by the success of Deep Learning, the scientific community proposed several approaches to make the learning procedure of Neural Networks more effective. 
When focussing on the way in which the training data are provided to the learning machine, we can distinguish between the classic random selection of stochastic gradient-based optimization and more involved techniques that devise curricula to organize data, and progressively increase the complexity of the training set. In this paper, we propose a novel training procedure named Friendly Training that, differently from the aforementioned approaches, involves altering the training examples in order to help the model to better fulfil its learning criterion. The model is allowed to ``simplify'' those examples that are too hard to be classified at a certain stage of the training procedure. The data transformation is controlled by a developmental plan that progressively reduces its impact during training, until it completely vanishes. In a sense, this is the opposite of what is commonly done in order to increase robustness against adversarial examples, i.e., Adversarial Training. Experiments on multiple datasets are provided, showing that Friendly Training yields improvements with respect to informed data sub-selection routines and random selection, especially in deep convolutional architectures. Results suggest that adapting the input data is a feasible way to stabilize learning and improve the generalization skills of the network.
\end{abstract}


\section{Introduction}
\label{sec:intro}

The outstanding results yielded by Neural Networks in several real-world tasks in the last decade \cite{alexnet,alphago,gpt3} have strongly motivated researchers to put even more effort in improving the design of neural architectures and to introduce tools to make the learning procedure more effective. In the framework of optimization-based learning, a number of improved methods are nowadays extremely popular across different application fields: regularization techniques \cite{dropout}, learnable normalization functions  \cite{batchnormalization}, powerful adaptive methods to estimate the appropriate learning rates \cite{adam}, to name a few. 

In this paper we propose an approach that introduces a novel perspective in the learning dynamics of Neural Networks, and is compatible with the aforementioned techniques. In particular, we consider the possibility of extending the search space of the learning algorithm, allowing the network not only to adapt its weights and biases, but also to adapt on-the-fly the training data, in order to facilitate the fulfilment of its learning criterion. Of course, such adaptation, that we also refer to as ``simplification'', must be controlled and embedded into a precise developmental plan in which the machine is progressively constrained to reduce the amount of simplification, until it ends up in handling the original training data as they are.  

\begin{figure*}[!ht]
\centering
\begin{minipage}{0.025\textwidth}
      \rotatebox{90}{\hspace{4.5mm} Friendly Training \hspace{1.05cm} Classic Training}
\end{minipage}
\begin{minipage}{0.85\textwidth}
\begin{multicols}{4}
\includegraphics[width=\linewidth]{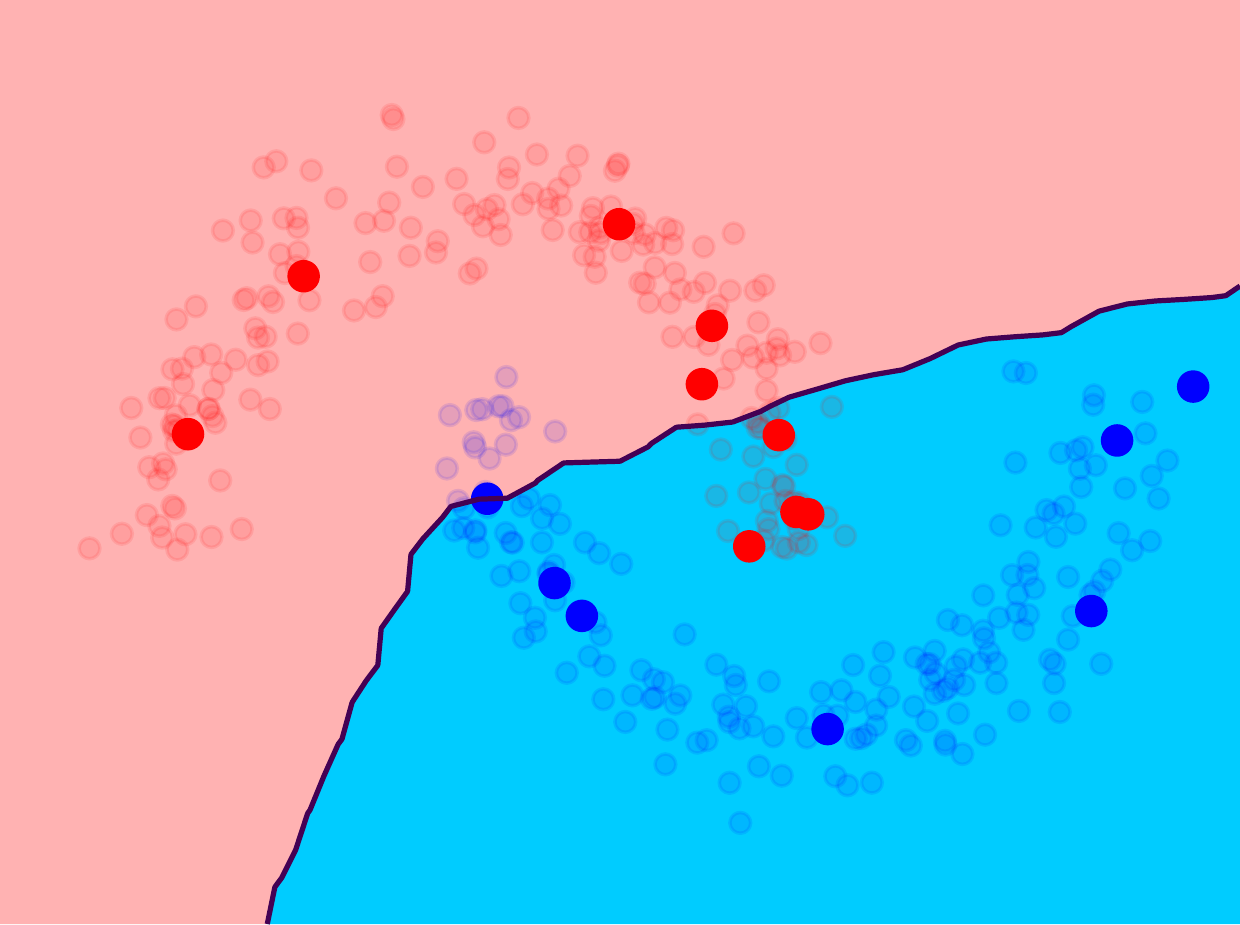} \par\caption*{Iteration $\gamma$ = 10}
    \includegraphics[width=\linewidth]{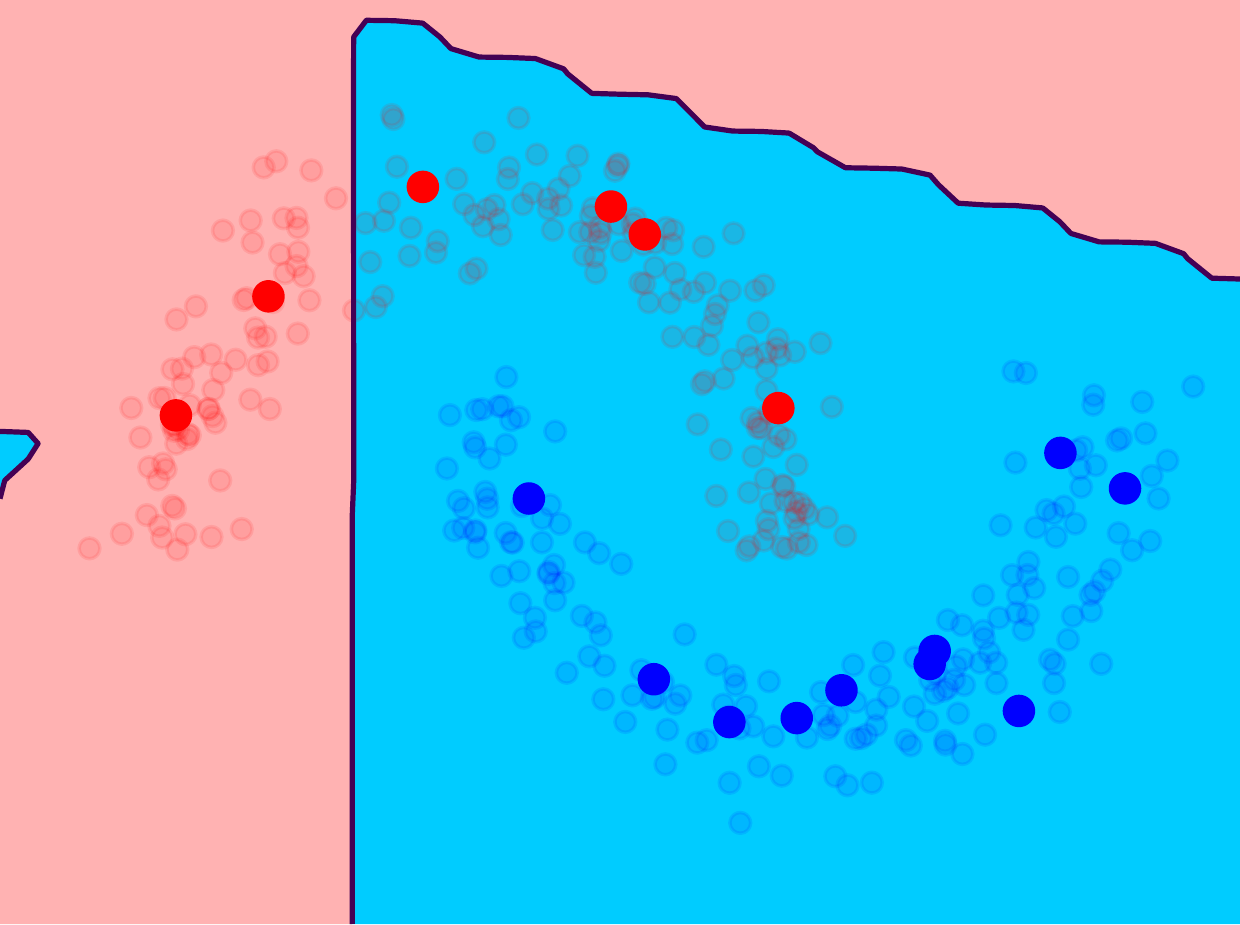}\par\caption*{Iteration $\gamma$ = 300}
    \includegraphics[width=\linewidth]{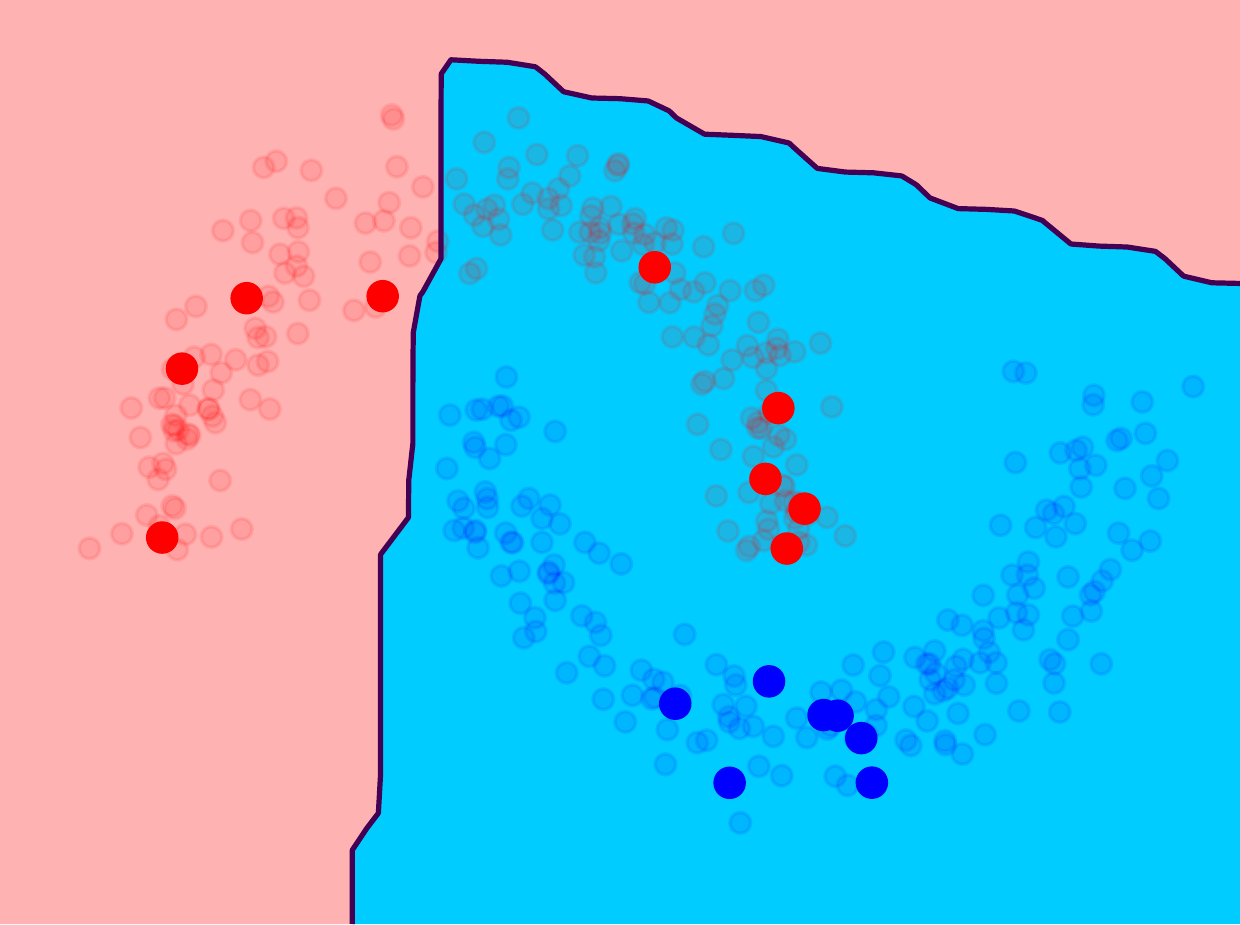}\par\caption*{Iteration $\gamma$ = 600}
    \includegraphics[width=\linewidth]{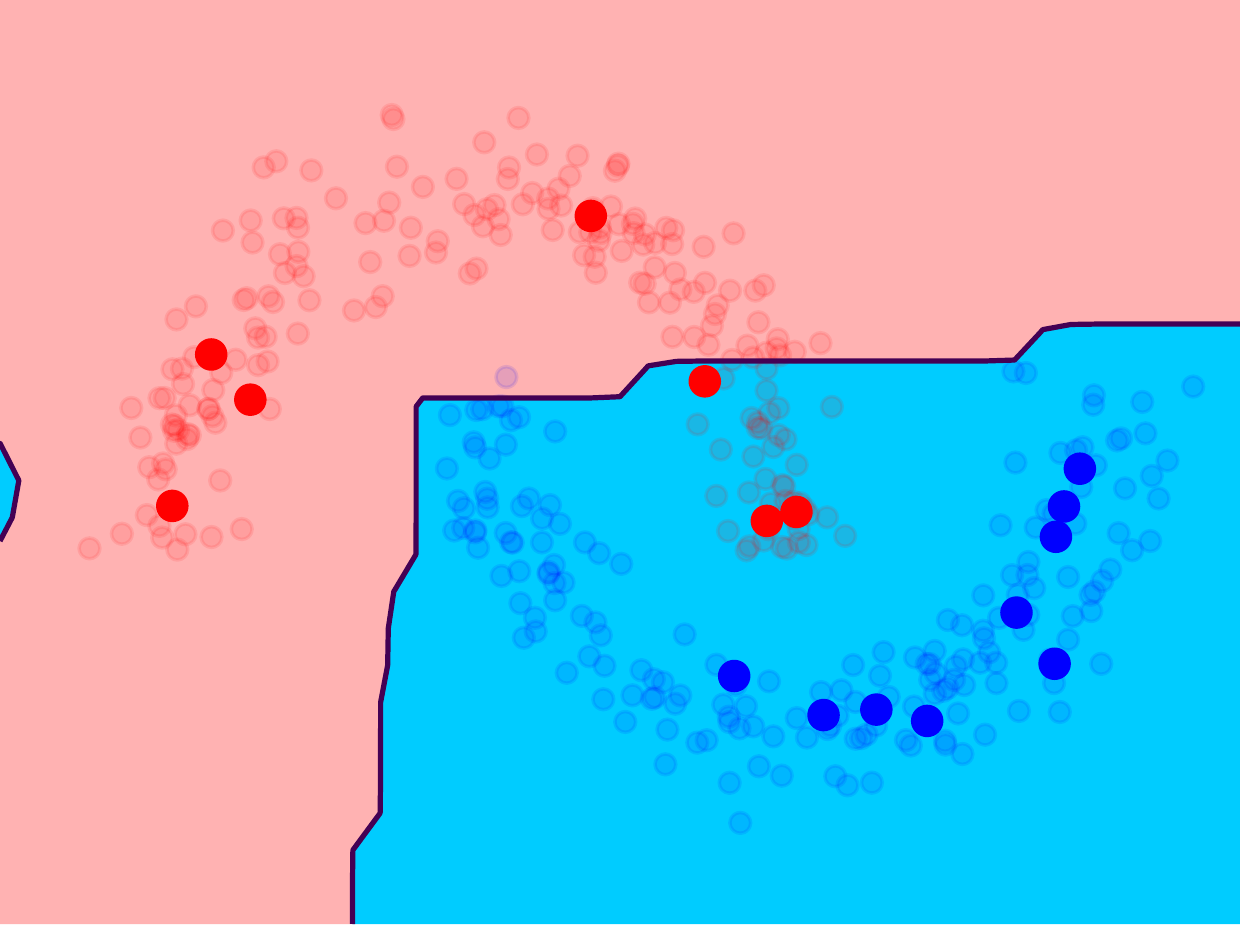}\par\caption*{Iteration $\gamma$ = 1000}
\end{multicols}
\begin{multicols}{4}
\includegraphics[width=\linewidth]{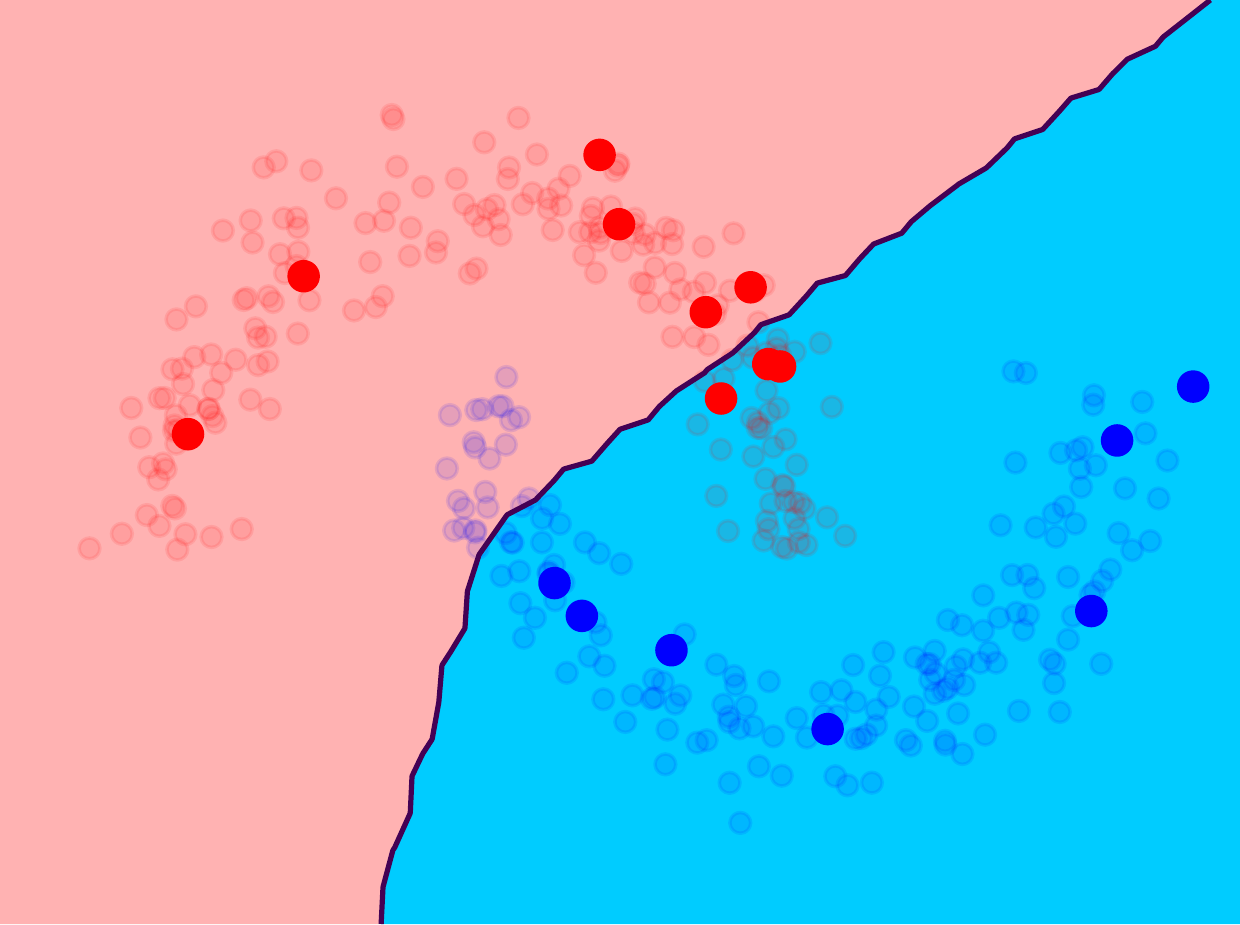}\par\caption*{Iteration $\gamma$ = 10}
    \includegraphics[width=\linewidth]{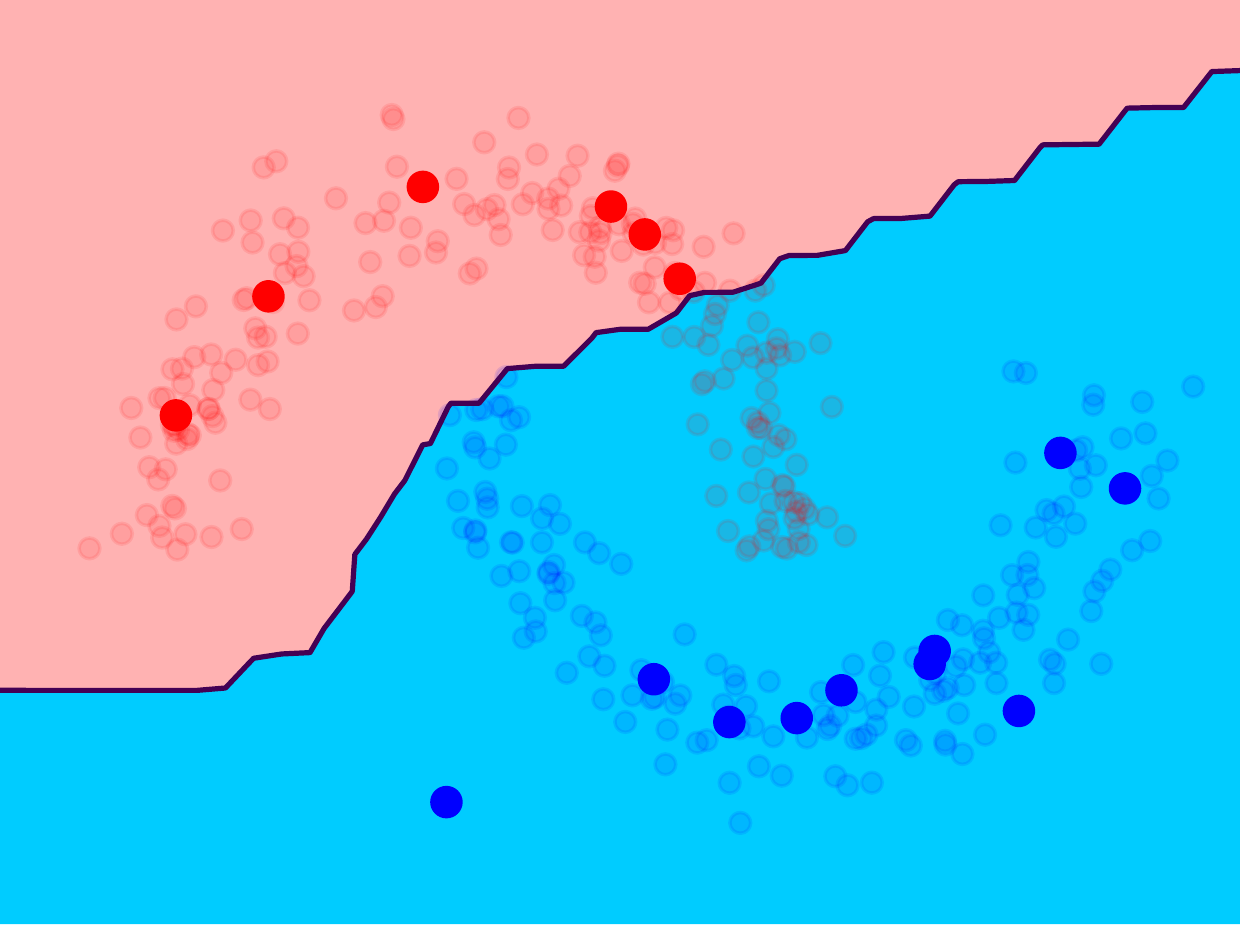}\par\caption*{Iteration $\gamma$ = 300}
    \includegraphics[width=\linewidth]{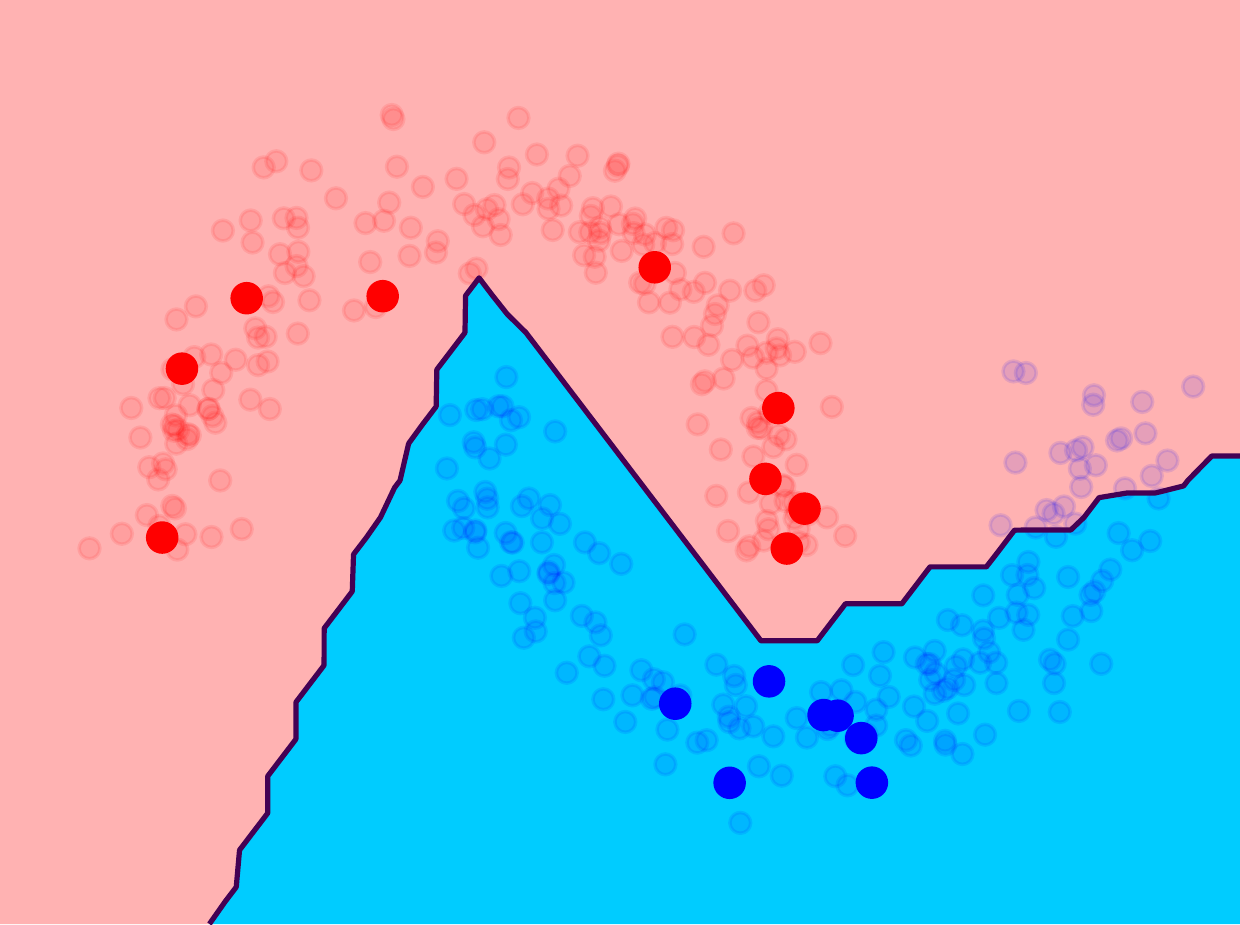}\par\caption*{Iteration $\gamma$ = 600}
    \includegraphics[width=\linewidth]{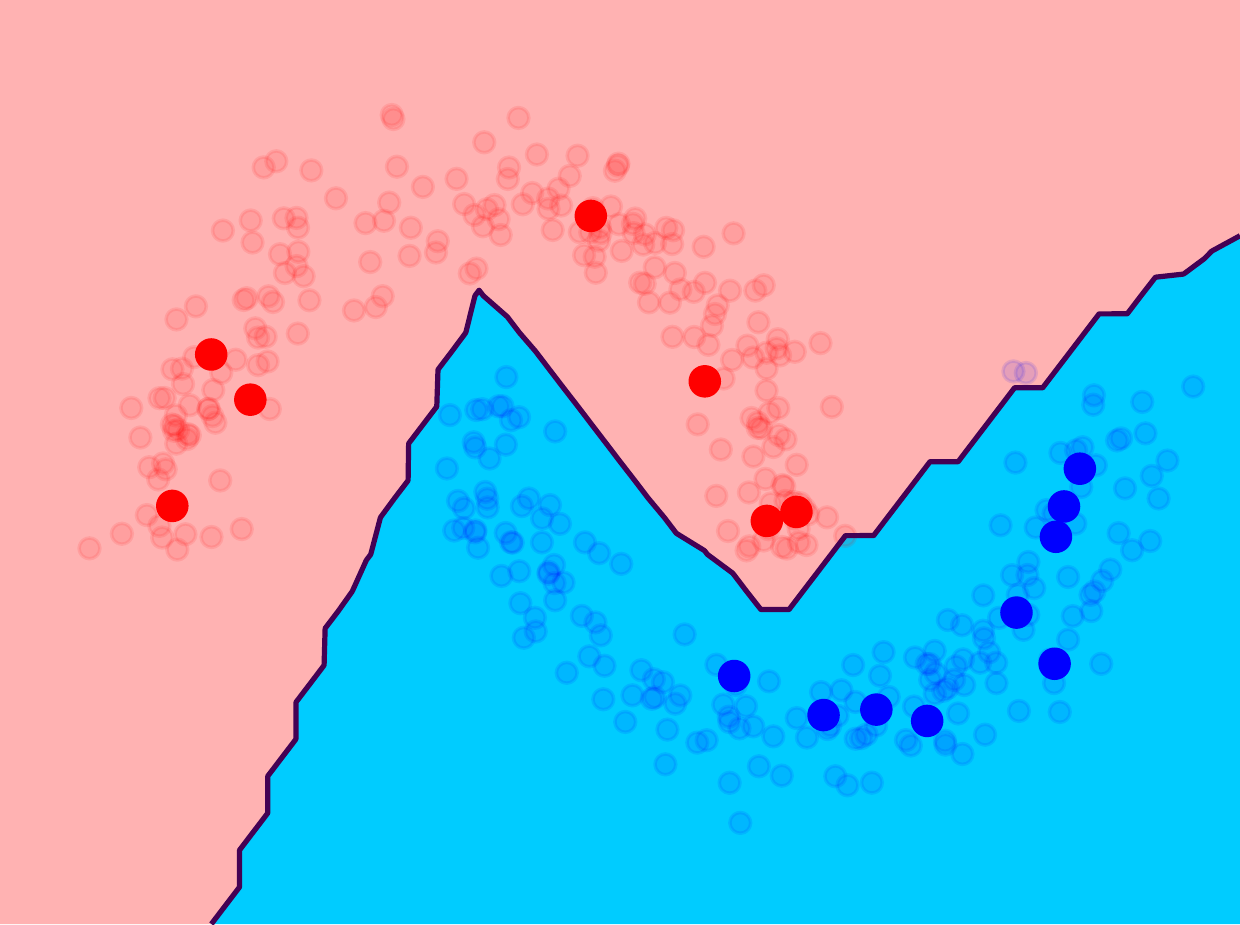}\par\caption*{Iteration $\gamma$ = 1000}
\end{multicols}
\end{minipage}
\caption{Temporal evolution of the decision boundary developed by a single hidden layer network (5 neurons, \texttt{tanh} activation, Adam optimizer) when learning on the two-moon dataset. Each plot is about a different training iteration ($\gamma$), and reports the full training set (semi-transparent empty circles) and the data of the last processed mini-batch of size $16$ (filled circles). Top row: classic training. Bottom row: Friendly Training. Both the experiments were initialized with the same weights and shared all the hyper-parameters. Those mini-batch examples (filled circles) that appear in different locations in the bottom-row with respect to the top-row have been temporarily altered by Friendly Training. As it can be noticed, the decision boundary adapts to the training data more smoothly and coherently when Friendly Training is employed.}
\label{fig:toy}
\end{figure*}

The key intuition behind this training strategy, that we named Friendly Training (FT), is that instead of exposing the network to an uncontrolled variety of data with heterogeneous properties over the input space, the learning can rather be guided by the information that the network has learnt to process so far. 
The aim of such technique is to operate on noisy data, outliers, and, more generally, on whatever falls into the areas of the input space that the network finds hard to handle at a certain stage of the learning process. Such data are modified to mitigate the impact of the information that is inconsistent with what has been learnt so far. 
In a sense, FT alters data so that they are just slightly more complicated with respect to what the learner is currently able to process, thus inducing the network to smoothly update the learnt function. This procedure is iterated until, at the end of training, the learnt function is consistent with the information carried by the original training data (see the toy example of Fig.~\ref{fig:toy}). This strategy echoes the pedagogical approach formulated by the psychologist Lev Vygotsky \cite{zsp}, according to which children learn by a progressive reduction of the so-called Zone of Proximal Development, \ie,  the space between what the learner can and cannot still do autonomously, which contains tasks that the learner can accomplish if appropriately guided.  
In our approach, the network itself decides how to adapt the data to foster its learning process, without the need of extra information on the problem or additional scoring functions.

The idea of providing data to the network following an easy-data-first plan has been popularized in the last decade by Curriculum Learning (CL) \cite{bengiocurriculum}. In this case, the training set is progressively expanded by adding more difficult data, that are selected and ordered accordingly to a scoring function that might come from further knowledge on the considered problem or other heuristics. A recent study \cite{Wu2020WhenDC} on the impact of CL showed the importance of CL when dealing with noisy settings or in case of limited time or computational resources. The idea of CL has been recently applied to the case of convolutional architectures with a progressive smoothing of the convolutional feature maps \cite{sinha-curriculum}. A related research area is the one of Self-Paced Learning \cite{spl-kumar}, in which some examples are either excluded from the training set or their impact in the loss function is kept low if some conditions, modeled by a specific regularizer, are met \cite{spcn}. Such conditions depend on the current state of the classifier, and the whole process is repeated multiple times. Differently from what we propose, these techniques do not alter the training data. From the algorithmic point of view, our approach can be seen as related to the so-called Adversarial Training strategies \cite{madry2019deep,zhang2020fat}, although  oriented toward completely different goals. 
Such strategies exploit the idea of providing the network with examples specifically generated to be potentially capable of fooling the classifier. Such samples are then added to the original training data to make the classifier more robust to adversarial attacks.
The price to pay is that this procedure might have a negative impact on the generalization skills of the classifier \cite{raghunathan2019adversarial}. In this respect, it has been recently shown that it is better to limit the data alteration during the early stages of the learning process, in order to implement a Friendly Adversarial Training policy \cite{zhang2020fat}. We partially borrowed this intuition, although focussing on the opposite direction, i.e., we let the classifier help itself by altering the data, instead of creating more difficult training conditions. 
The aim of our approach is to improve the network generalization capability by letting the network learn in a friendly environment.

In this manuscript, we provide an experimental analysis conducted on five datasets available in the related literature. Results suggest that Friendly Training is a feasible way to improve the generalization skills of the network, and to create a smooth evolution of the learning environment, avoiding to expose the network to uncontrolled information at the wrong time. The simplifications, on average, look more structured in (deep) convolutional architectures, leading to more effective improvements, while their impact in fully-connected networks is more spread over all the dimensions of the input space, resulting in less appreciable gains. In the following, we provide an in-depth analysis of the sensitivity of the system to the newly introduced hyper-parameters, and we compare Friendly Training with randomly ordered data or curricula defined by ad-hoc criteria.

The contributions of this paper are the following ones: (1) we propose a novel training strategy, named Friendly Training (FT), that allows the machine to partially simplify the data by automatically determining how to alter it; (2) we propose a developmental plan that allows the effects of FT to progressively fade out, in order to create a smooth transition from simplified data to the original one; (3) we experimentally evaluate the FT approach in convolutional and fully connected neural architectures with different number of layers and considering five different datasets, analyzing the impact of the simplification.

This paper is organized as follows. FT, including the developmental plan, is described in Section~\ref{sec:method}. The relation with related work are deeply analyzed in Section~\ref{sec:related}. Experiments and results are described in Section~\ref{sec:methods}, while conclusions and suggestions for future work are drawn in Section~\ref{sec:conclusions}.

\section{Friendly Training}
\label{sec:method}

We consider a generic classification problem in which we are given a training set $\mathcal{X}$ composed of $n$ supervised pairs, $\mathcal{X} = \{ (x_k, y_k),\ k=1,\ldots,n \}$, being $x_k \in \mathbb{R}^{d}$ a training example labeled with $y_k$.\footnote{We consider the case of classification mostly for the sake of simplicity. The proposed approach goes beyond classification problems.}
Given some input data $x$, we denote with $f(x, w)$ the function computed by a Neural Network-based classifier with all its weights and biases stored into vector $w$. When optimizing the model exploiting a mini-batch based stochastic gradient descent procedure, at each step of the training routine the following loss function $L$ is used for computing the gradient with respect to $w$ and updating the model parameters:
\begin{equation}
    L\left(\mathcal{B}, w \right) = \frac{1}{|\mathcal{B}|} \sum_{i=1}^{|\mathcal{B}|} \ell \left(f\left(x_{i}, w \right) , y_i\right), 
    \label{eq:loss}
\end{equation}
where $\mathcal{B} \subset \mathcal{X}$ is a mini-batch of data of size $|\mathcal{B}| \geq 1$, $(x_i, y_i) \in \mathcal{B}$, and $\ell$ is the loss restricted to a single example. For simplicity, we assumed to aggregate the contributes of $\ell$ by averaging over the mini-batch data. The set $\mathcal{B}$ can be sampled with different strategies. In the most common case of stochastic gradient optimization, a set of mini-batches is randomly sampled at each training epoch, in order to cover the whole set $\mathcal{X}$ without intersections among the mini-batches.

Of course, the data that populate $\mathcal{X}$ might include examples with different properties, in a task-dependent manner. To give some examples, the distribution of the training data could be multi-modal, might include outliers, might span over several disjoint manifolds, and so on. However, the aforementioned training procedure provides data to the machine independently on the state of the network and with no control on the information carried by such data. If additional information is available on the problem at hand, it might be exploited to devise curricula providing simple-examples-first, as in CL \cite{bengiocurriculum}. However, it is very unlikely to have information on the complexity of examples in advance and, even more importantly, the human-based criteria might not match the way in which examples are processed by the machine. The value $\ell$ could be used as an indicator to estimate such simplicity, to exclude those examples with too large loss or to reduce their contribution in Eq.~(\ref{eq:loss}), similarly to what is done in \cite{spl-kumar,spcn}.\footnote{We postpone to Section~\ref{sec:related} a more detailed description of these strategies.}

In this paper, instead, we propose an alternative approach that, differently from what has been described so far, analyzes the training data according to the state of the learner, allowing the network to modify such data, eventually discarding the parts of information that are too complex to be handled by the model at that moment, while preserving what sounds more coherent with the expectation of the current classifier. Notice that this is significantly different from deciding whether or not to keep a training example, to give small or big weight to them in Eq.~(\ref{eq:loss}), or to simply re-order the examples. Interestingly, FT is compatible with (and not necessarily an alternative to) the aforementioned existing strategies.

Formally, each example $x_i$ is altered to $\tilde{x}_i$ by adding a learnable perturbation $\delta_i$,
\begin{equation}
    \tilde{x}_i = x_i + \delta_i,
    \label{eq:delta}
\end{equation}
where $\delta_i \in \mathbb{R}^{d}$. Given a mini-batch $\mathcal{B}$, we indicate with $\Delta$ the matrix that collects the perturbations associated to the mini-batch examples. In detail, the $i$-th row of $\Delta$ is the perturbation $\delta_i$ associated with the $i$-th example in $\mathcal{B}$.
For convenience in the notation, we avoid mentioning training epochs in what follows, and we describe the training procedure as the iterative processing of mini-batches of data, updating $w$ after each of them. Let us denote with $\gamma \geq 1$ the iteration index.
We re-define the aggregated loss $L$ of Eq.~(\ref{eq:loss}) by providing the network with $\tilde{x}$ instead of $x$, introducing the dependency on $\Delta$, and by adding the iteration index $\gamma$,
\begin{equation}
    L(\mathcal{B}^{\gamma},\Delta^{\gamma},w^{\gamma}) = \frac{1}{|\mathcal{B}^{\gamma}|} \sum_{i=1}^{|\mathcal{B}^{\gamma}|} \ell \left(f\left(\tilde{x}_{i},w^{\gamma} \right) , y_i\right), 
    \label{eq:loss2}
\end{equation}
where $\tilde{x}_i$ is defined as in Eq.~(\ref{eq:delta}), $(x_i,y_i) \in \mathcal{B}^{\gamma}$ (i.e., $\mathcal{B}^{\gamma}$ is the mini-batch at iteration $\gamma$) and $\delta_i$ is the $i$-th row of $\Delta^{\gamma}$. 
Jointly optimizing Eq.~(\ref{eq:loss2}) with respect to $\Delta^{\gamma}$ and $w^{\gamma}$ allows the network not only to adapt its weights and biases in order to better cope with the learning criterion, but also to alter the data in $\mathcal{B}^{\gamma}$ by translating them in those space regions that can be more easily classified. However, the loss of Eq.~(\ref{eq:loss2}) does not introduce any constraints on each $\delta_i$. Hence,  the network is free to change the training data without any guarantees that the simplification amount will reduce while learning proceeds.
Moreover, differently from $w^{\gamma}$, the set $\Delta^{\gamma}$ is specifically associated with the data in $\mathcal{B}^{\gamma}$ (i.e, each training example is associated with its own perturbation), meaning that the number of variables of the optimization problem becomes a function of the size of the training data.

We frame Eq.~(\ref{eq:loss2}) in the context of a developmental plan that solves both these issues. First, the system is enforced to reduce the perturbations as long as the number of training iterations increases. If $\gamma_{max}$ is the maximum number of allowed iterations, we ensure that after $\gamma_{max\_simp} < \gamma_{max}$ steps the data are not perturbed anymore.
Secondly, we remove the dependence of $\Delta^{\gamma}$ from $\gamma$, introducing an Alternate Optimization scheme in which we decouple the optimization of perturbations and weights.
In detail, we consider a single matrix $\Delta$ that is shared by all the mini-batches (the total number of rows in $\Delta$ is equal to the size of a single mini-batch). At the beginning of each training iteration, we keep $w^{\gamma}$ fixed, we initialize $\Delta$ to zeros and then we estimate the appropriate perturbations for the current data in $\mathcal{B}^{\gamma}$ by gradient descent over the variable $\Delta$ (second argument of Eq.~(\ref{eq:loss2})). We indicate with $\tau^{\gamma}$ the number of iterative steps of such inner optimization. The value of $\tau^{\gamma}$ controls the amount of alteration on the data. For small values of $\tau^{\gamma}$ the network will only marginally simplify the data, while for a larger $\tau^{\gamma}$ the data alteration will be more aggressive.
The initial value $\tau^{1} \geq 1$ is a fixed hyper-parameter, while we considered a quadratic law to progressively reduce $\tau$ in function of $\gamma$,
\begin{equation}
    \tau^{\gamma} = \tau^{1} \cdot \max \left( 1 - \frac{\gamma - 1}{\gamma_{max\_simp}}, 0 \right)^2.
    \label{tauplan}
\end{equation}
Afterwards, we update the values of $w^{\gamma}$, given the output $\Delta$ at the end of the just described inner optimization. This developmental plan allows the system to adapt the data in order to make the learning procedure less disruptive, especially during the early stages of learning, introducing a smooth optimization path driven by the evolution of $\tau^{\gamma}$.
The detailed training procedure is formally reported in Algorithm~\ref{alg:friendly},
\begin{algorithm}
 \caption{Friendly Training.}
 \begin{algorithmic}[1]
 \renewcommand{\algorithmicrequire}{\textbf{Input:}}
 \renewcommand{\algorithmicensure}{\textbf{Output:}}
 \REQUIRE Training set $\mathcal{X}$, initial weights and biases $w^{1}$, batch size $b$, max learning steps $\gamma_{max}$, max simplification steps $\gamma_{max\_simp}$, $\tau^{1} \geq 1$, learning rates $\alpha > 0$ and $\eta > 0$, shared matrix $\Delta$ with $b$ rows and $d$ columns.
 \ENSURE The final $w^{\gamma_{max} + 1}$.
  \FOR {$\gamma = 1$ to $\gamma_{max}$}
  \STATE Sample $\mathcal{B}^{\gamma}$ of size $b$ from $\mathcal{X}$
  \STATE Compute $\tau^{\gamma}$ following Eq.~(\ref{tauplan}).
  \STATE Set all the entries of $\Delta$ to 0
  \FOR {$\tau = 1$ to $\tau^{\gamma}$}
  \STATE Compute $\Delta_{grad} = \frac{\partial L(\mathcal{B}^{\gamma},D,w^{\gamma})}{\partial D}\Bigr|_{\substack{D=\Delta}}$, see Eq.~(\ref{eq:loss2})
  \STATE $\Delta = \Delta - \eta \cdot \Delta_{grad}$
  \ENDFOR
    \STATE Compute $w^{\gamma}_{grad} = \frac{\partial L(\mathcal{B}^{\gamma},\Delta,v)}{\partial v}\Bigr|_{\substack{v=w^{\gamma}}}$, see Eq.~(\ref{eq:loss2})
  \STATE $w^{\gamma+1} = w^{\gamma} - \alpha \cdot w^{\gamma}_{grad}$  
  \ENDFOR
 \RETURN $w^{\gamma_{max} + 1}$ 
 \end{algorithmic}
 \label{alg:friendly}
 \end{algorithm}
and in the following lines we provide some further details. Notice that while the weight update equation (line 10) can include any existing adaptive learning rate estimation procedure, in our current implementation $\Delta$ is updated by a fixed small learning rate (line 7). While Algorithm~\ref{alg:friendly} formally returns the weights after having completed the last training iteration, as usual, the best configuration of the classifier can be selected by measuring the performance on a validation set, when available. Another important fact to mention is that when the prediction on examples perfectly matches target, line 6 will return zero gradient, hence no simplifications are performed. In our implementation we slightly relaxed this condition by zeroing the rows of $\Delta_{grad}$ (line 6) associated to those examples that are classified with a large confidence above a given threshold (see Section~\ref{sec:methods}), that implements a selective early-stopping condition on the inner optimization.

We qualitatively show the behavior of the proposed training strategy in the toy example of Fig.~\ref{fig:toy}. A very simple network with one hidden layer ($5$ neurons with hyperbolic tangent activation function) is trained on the popular two-moon dataset (two data distributions shaped as interleaving half-circles), optimized by Adam with mini-batch of size $16$. When the network (having the same initial weights) is trained with the Friendly Training algorithm, it is less subject to oscillations in the learning process, resulting in a more controlled development of the classifier and leading to a final decision boundary that better fits the data distribution. With the Friendly Training approach, the mini-batch examples (filled dots) are altered during the first iterations (second row, first two pictures - compare their position with the corresponding pictures in the first row, which represent Classic Training). Notice that the mini-batch examples occupy their original positions during later stages (final stage of developmental plan).

\section{Relationships with Existing Work}
\label{sec:related}

The idea of training neural networks with a learning methodology that ``gradually'' changes the learning environment (which proves to be appropriate not only in humans but also in animals \cite{shaping}) traces back to almost three decades ago \cite{elman} and has been known for long as Curriculum Learning (CL) \cite{bengiocurriculum}. 
Taking inspiration from the common experience of learning in humans, CL aims at designing an optimal learning plan in which the learning agent is exposed to simple, easily-discernible examples at first, and later to gradually harder examples, also progressively increasing the size of the training set.
CL proved to be successful in training deep networks on a variety of tasks \cite{gong2016multi, hacohen2019power}. 
The concrete effect of CL has been recently revisited \cite{Wu2020WhenDC}, showing how the positive impact of curriculum-based learning strategies is mostly evident in noisy setups (\eg, random permutation of labels) or large-data regimes with limited time availability. 
Nonetheless, recent evidence suggests that elementary curriculum-based tricks might help in vision-related tasks, at least when very deep and wide networks are used. Curriculum By Smoothing (CBS) \cite{sinha-curriculum} is built upon the idea of applying Gaussian-based low-pass filters on feature maps of Convolutional Neural Networks (CNNs) that process image data, with variance that progressively goes to zero as training proceeds.
Even though CL remains a somewhat controversial topic and has not been widely adopted in the machine learning community; more research may be needed before ruling out this approach. In fact, there is a quite general consensus on the fact that what happens in the early stages of deep network training affects the network behavior at steady state \cite{jastrzebski2020break}. For instance, the authors of \cite{achille2018critical} showed that inflicting visual impairments to deep CNNs in the first epochs leads to minor visual skills despite conceding recovery time. 

The purpose of Friendly Training, on the opposite, is to downplay difficult, confusing and outlier examples at the beginning, while let them contribute to the generalization capability when the learner has already acquired basic skills.
With respect to CL, Friendly Training does not alter the size of the dataset as a function of the number of training iterations, nor the relative ordering in which examples are presented to the network. Moreover it does not assume any predefined complexity criteria. Conversely, Friendly Training alters every single example in an adaptive way, unless is it correctly predicted with a sufficiently high confidence, and only afterwards the example is used to update the network weights. In so doing, the network is exposed, at every training step, to a data population with quite large variability, although their distribution has been slightly adapted, in order to decrease the occurrence of abrupt weight changes. Differently from CBS, Friendly Training can be applied to any type of data and, interestingly, it is compatible with CBS, and not necessarily an alternative to it.

Self-Paced Learning (SPL) \cite{spl-kumar} is a technique inspired by CL, originally designed for Structural Support Vector Machines (SSVMs) with latent variables. Self-paced stands for the fact that the curriculum is determined by the pupil’s abilities (the classifier's behaviour) rather than by a teacher's plan. Since this direction proved to be effective when compared with standard algorithms (latent variable models correspond to hard optimization problems), researchers adapted this formulation to CNNs \cite{spcn}. The basic idea consists of searching for suitable example-specific weighting coefficients in the loss computation.
In contrast to this approach, what we propose is not about weighting the importance of training examples, but rather temporarily altering them. 
However, we do embrace the idea that the state of the classifier is what can be used to determine how to deal with a particular input/target pair, gradually exposing the learner to more and more difficult examples, with an explicit temporal dynamics.


Another technique that is somehow related to Friendly Training is the so-called Adversarial Training (AT) \cite{goodfellow-adversarial,madry2019deep}, which may be seen as the inverse learning technique of FT. Developed as an empirical defense strategy for adversarial attacks, which seriously affect neural networks operating on high-dimensional spaces \cite{goodfellow-adversarial}, AT incorporates adversarial data into the training process. Most AT techniques rely on a minimax optimization problem \cite{madry2019deep}, since the goal is to generate adversarial examples that strongly fool the classifier. Each generated example is an artificial element, lying very close to an original training data point of a certain class, but that is classified incoherently with respect to the label attached to such original point. Interestingly, in AT the system basically alters examples as in Eq.~(\ref{eq:delta}), even if the perturbation is computed with a different criterion and typically with no temporal dynamics.

Friendly Adversarial Training (FAT) \cite{zhang2020fat} builds up on the ideas of both CL and AT. Researchers noticed \cite{tsipras2019robustness} that the adversarial formulation sometimes hurts generalization capabilities \cite{raghunathan2019adversarial}. The FAT strategy provides a more gentle learning problem, in which the generation of adversarial data is early-stopped as soon as the datapoint is misclassified. The resulting learning dynamics is such that, as learning progresses (together with accuracy and robustness), more and more iterations are needed to generate (harder) adversarial data. Our Friendly Training algorithm shares some intuitions with the FAT algorithm (consider the comments on early-stopping right after Algorithm~\ref{alg:friendly}), even though the direction of the iterative process which generates input data has a different goal. Moreover, FAT does not include explicit temporal dynamics.

\section{Experiments}
\label{sec:methods}

We describe our experimental experience by introducing the considered datasets in Section~\ref{sec:data}, the neural architectures, the parameter tuning procedure and the competitors in Section~\ref{sec:setup}, and by reporting and discussing the numerical and qualitative results in Section~\ref{sec:results}, also including an analysis of the sensitivity of key parameters.
\subsection{Datasets}
\label{sec:data}

In order to assess the performance of the proposed learning algorithm, we employed the datasets presented in \cite{mnistvariations}. Some of these datasets  are about $10$-class digit recognition problems ($28\times 28$, grayscale), and they were explicitly designed to provide harder learning conditions with respect to the well-known MNIST dataset, keeping an affordable size (62k examples per dataset, already divided into training, validation and test set). The data distribution is the product of multiple factor distributions (\ie, rotation angle, background, etc., besides factors inherent to the original data, such as the handwriting style), making them a challenging benchmark. In detail:
\begin{itemize}
\item \emph{mnist-rot}: MNIST digits, rotated by a random angle $\theta$, so that $\theta \in [0, 2\pi]$.
\item \emph{mnist-back-image}: MNIST digits, with the background replaced with patches extracted by random images  of public domain.
\item \emph{mnist-rot-back-image}: MNIST digits, with both the rotation and background factors of variations (combined).
\end{itemize}
We also considered other different types of data still from \cite{mnistvariations}, aimed at investigating how neural networks deal with geometrical shapes and learn their properties. They share the same resolution and almost the same size of the previously mentioned datasets, but they are used for geometry-based binary classification tasks. They are:
\begin{itemize}
\item \emph{rectangles-image}: white rectangles, wide or tall, with the inner and outer regions filled with patches taken from random images.
\item \emph{convex}: convex or non-convex white regions on a black background.
\end{itemize}

\subsection{Experimental Setup}
\label{sec:setup}
We focused on four neural network architectures, where two of them are feed-forward Fully-Connected multi-layer perceptrons, referred to as FC-A and FC-B, and the others are Convolutional Neural Networks named CNN-A and CNN-B. FC-A is a simple one-hidden-layer network with hyperbolic tangent activations (10 hidden neurons), while FC-B is inherited from \cite{achille2018critical}, has 5 hidden layers (2500-2000-1500-1000-500 neurons), batch normalization and ReLU activations.  CNN-A consists of 2 convolutional layers, max pooling, dropout and 2 fully connected layers; CNN-B is deeper (4 convolutional layers). Both of them employ ReLU activation functions on the convolutional feature maps (32-64 filters in CNN-A, 32-48-64-64 filters in CNN-B) and on the fully connected layers activations (9216-128 neurons for CNN-A, 5184-128 neurons for CNN-B). Cross-entropy is used as loss function, while the Adam optimizer is exploited for the network weights and biases, using mini-batches of size $32$.

We compared the test error rates of these architectures when trained under the following conditions:
\begin{itemize}
\item \textit{Classic Training (CT)}: training of the neural network parameters with random selection of examples (no intervention on data), as in most nowadays cases.
\item \textit{Friendly Training (FT)}: the FT training procedure of Algorithm~\ref{alg:friendly} is exploited.
\item \textit{Easy-Examples First (EEF)}: mini-batch examples are sorted by loss value in ascending order and only the first $k$ of them are used to calculate the gradients. At the beginning, $k=1$, so that only $1$ example per mini-batch is kept, then $k$ grows with the training iterations following the same dynamics of FT. The criterion implemented by EEF can be identified as a basic instance of CL, closely related to FT (i.e., it is has the same temporal dynamics of FT, but examples are not altered and only sub-selected).
\end{itemize}
All the experiments are executed for 200 epochs (trivially, $\gamma_{max}$ is the number of epochs multiplied by the number of mini-batches per epoch), which consistently proved to be sufficient to obtain convergence in all the learning problems. Moreover, the metrics we report are about the epoch with the lowest validation error. All the experiments referring to the same architecture share the same initialization weights; examples are presented in the same order.

The hyper-parameters of FT and EEF were selected by grid search. We indicate with $c$ the threshold above which a prediction is considered correct in order to early stop the data transformation of FT (see the description right after Algorithm~\ref{alg:friendly}). In detail, we considered the following ranges: $\eta \in [0.01, 0.1, 1.0, 5.0, 10.0]$, $\tau^{1} \in [10, 80, 120]$, $c \in [0.9, 0.95, 0.98]$, $\gamma_{max\_simp} \in [0.25, 0.5, 0.85] \cdot \gamma_{max}$. The above intervals proved to be suitable in preliminary investigations. Each experiment was repeated 3 times (results are averaged), varying the initialization of the weights and biases (sharing them among the competitors).

\subsection{Results and Discussion}
\label{sec:results}

We report the test error rate we obtained in Table~\ref{tab:main-table}, along with some reference results by other authors \cite{mnistvariations,spcn} (top-portion of Table~\ref{tab:main-table}). Such reference results are about Fully-connected networks (NNet), Deep Belief Networks (DBNs, two different architectures), Convolutional Autoencoders (CAEs, two different architectures) -- see \cite{mnistvariations} for further details. General Stochastic Network (GSN) is described in \cite{zohrer2014general}. Finally, SPCN is an instance of Self-Paced Learning based on CNNs, described in \cite{spcn}. 

Regardless of the learning algorithm, convolutional architectures CNN-A and CNN-B consistently achieve lower error rates, as expected. Several of the reference results are beaten by the architectures we experimented, also in the case of CT, due to the fact that we considered models with a larger number of parameters. Nonetheless, the general trend is coherent with reference experiments, confirming the validity of our experimental setup (e.g., compare FC-A and FC-B with NNet, or SPCN with CNN-A and CNN-B, for example). 

When comparing \emph{CT}, \emph{EEF}, and \emph{FT}, we notice that albeit EEF occasionally improves the baseline (CT) result (\eg, in the mnist-back-image task with convolutional networks), it degrades the performance in most of the datasets, confirming that simply keeping the low-loss examples of the training set, although injected in a progressive developmental plan, does not work as a trivial trick to improve performance.
Differently, the FT algorithm clearly improves the test error rate when used on the convolutional architectures CNN-A and CNN-B, providing better results than the competitors in most of the cases. In the case of CNN-A, the error is systematically lowered (sometimes in a statistically significant manner), with the exception of the convex dataset. Regarding CNN-B, we still get improvements in four out of five datasets, even if not as evident as in CNN-A. These results confirm the validity of the proposed learning technique. Fully-Connected networks FC-A and FC-B, conversely, are usually not improved by FT, with a few exceptions. A further inspection on the way the system is computing the perturbation offsets revealed important elements that explain these results, pointing out at interesting facets of FT that we will describe in the following.

\begin{table*}[h]
\caption{Performance comparison of classifiers with different architectures (FC-A, FC-B, CNN-A, CNN-B) and learning algorithms (CT, EEF, FT). Mean test error (smaller is better) is reported along with standard deviation. The first portion of rows is about reference results taken from existing work (see the paper text for more details). For each architecture, we report in bold those results that improve the baseline (CT) case.}
\label{tab:main-table}
\begin{center}
  \begin{tabular}{c|cccHcc}
     \toprule
      Classifier&mnist-back-image&mnist-rot-back-image&mnist-rot&rectangles&rectangles-image&convex\\ \midrule
NNet&$27.41 \pm 0.39$&$62.16 \pm 0.43$&$18.11 \pm 0.34$&$7.16 \pm 0.23$&$33.20 \pm 0.41$&$32.25 \pm 0.41$\\
DBN-1&$16.15 \pm 0.32$&$52.21 \pm 0.44$&$14.69 \pm 0.31$&$4.71 \pm 0.19$&$23.69 \pm 0.37$&$19.92 \pm 0.35$\\
DBN-3&$16.31 \pm 0.32$&$47.39 \pm 0.44$&$10.30 \pm 0.27$&$2.60 \pm 0.14$&$22.50 \pm 0.37$&$18.63 \pm 0.34$\\
CAE-1&$16.70 \pm 0.33$&$48.10 \pm 0.44$&$11.59 \pm 0.28$&$1.48 \pm 0.10$&$21.86 \pm 0.36$& n.a.\\
CAE-2&$15.50 \pm 0.32$&$45.23 \pm 0.44$&$9.66 \pm 0.26$&$1.21 \pm 0.10$&$21.54 \pm 0.36$&n.a.\\
GSN&$16.04 \pm 0.07$&$43.86 \pm 0.05$&$8.66 \pm 0.08$&$2.04 \pm 0.04$&$22.10 \pm 0.03$&n.a.\\
SPCN&$9.55 \pm 0.06$&$35.26 \pm 0.05$&$9.81 \pm 0.07$&$0.19 \pm 0.03$&$10.60 \pm 0.03$&n.a.\\
\midrule
\midrule
FC-A / CT&$28.34 \pm 0.09$&${64.06} \pm 0.31$&${43.16} \pm 0.51$&$\textbf{17.34} \pm 1.85$&${24.31} \pm 0.21$&$33.91 \pm 0.44$\\
FC-A / EEF&$\textbf{28.18} \pm 0.47$&$64.27 \pm 0.19$&$43.91 \pm 0.73$&$23.92 \pm 1.58$&$24.48 \pm 0.11$&$\textbf{33.17} \pm 0.93$\\
FC-A / FT&$28.66 \pm 0.06$&$64.14 \pm 0.36$&$43.24 \pm 0.43$&$21.50 \pm 3.71$&$24.64 \pm 0.37$&$34.38 \pm 0.22$\\
\midrule 
FC-B / CT&${21.06} \pm 0.39$&$51.71 \pm 0.79$&$10.13 \pm 0.27$&$\textbf{4.86} \pm 0.31$&$25.10 \pm 0.20$&${27.24} \pm 0.05$\\
FC-B / EEF&$21.38 \pm 0.18$&$52.95 \pm 0.63$&$\textbf{10.04} \pm 0.17$&$6.13 \pm 0.78$&$\textbf{24.84} \pm 0.32$&$28.21 \pm 0.96$\\
FC-B / FT&$21.74 \pm 0.26$&$\textbf{51.02} \pm 0.07$&$11.19 \pm 0.37$&$5.35 \pm 0.80$&$\textbf{24.14} \pm 0.53$&$27.49 \pm 0.07$ \\
\midrule
CNN-A / CT &$7.25 \pm 0.16$&$29.05 \pm 0.45$&$7.48 \pm 0.14$&$0.04 \pm 0.01$&$9.86 \pm 0.32$&${8.24} \pm 0.09$\\
CNN-A / EEF&$\textbf{7.02} \pm 0.08$&$29.12 \pm 0.34$&$7.61 \pm 0.22$&$\textbf{0.03} \pm 0.01$&$12.82 \pm 0.70$&$8.72 \pm 0.74$\\
CNN-A / FT&$\textbf{6.80} \pm 0.19$&$\textbf{28.74} \pm 0.29$&$\textbf{7.36} \pm 0.06$&$0.04 \pm 0.01$&$\textbf{9.72} \pm 0.20$&$8.59 \pm 1.44$\\
\midrule
CNN-B / CT&$5.15 \pm 0.15$&$23.05 \pm 0.21$&$ {6.58} \pm {0.06} $&$ \textbf{0.02} \pm \textbf{0.01}$&$8.10 \pm 1.90$&$3.01 \pm 0.41$\\
CNN-B / EEF&$ \textbf{4.82} \pm {0.19}$&$\textbf{22.89} \pm 0.49$&$7.02 \pm 0.28$&$0.27 \pm 0.23$&$8.35 \pm 1.01$&$3.75 \pm 0.58$\\
CNN-B / FT&$\textbf{5.03} \pm 0.11$&$ \textbf{22.81} \pm {0.36}$&$6.95 \pm 0.12$&$0.18 \pm 0.09$&$ \textbf{7.32} \pm {1.31} $&$ \textbf{2.87} \pm {0.42}$\\

    \bottomrule
  \end{tabular}
\end{center}
\end{table*}

\begin{figure}[h]
\centering
\includegraphics[width=0.4\textwidth]{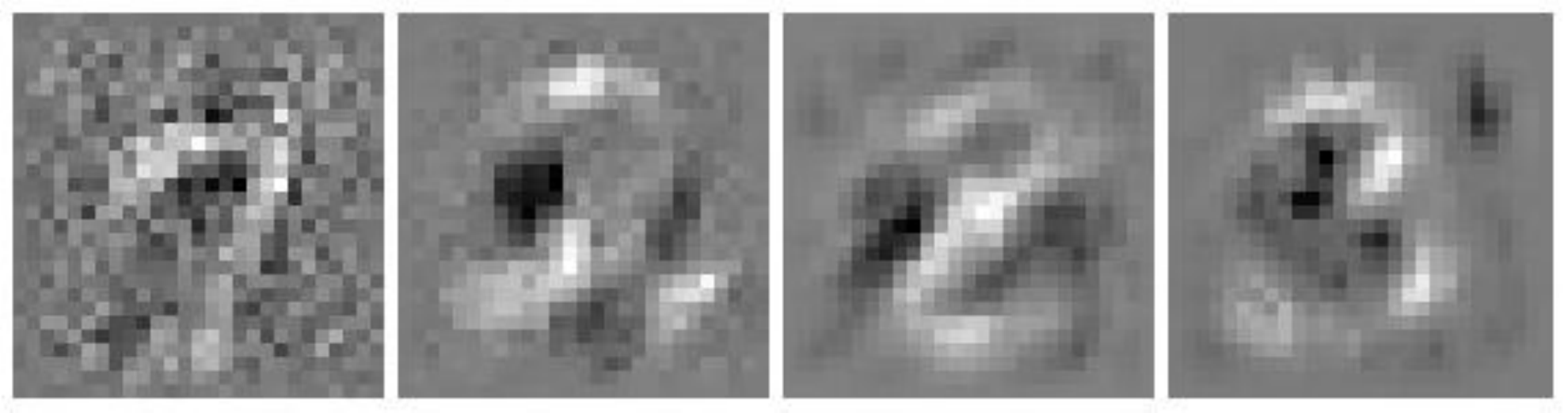}
\caption{Randomly selected perturbation offsets ($\delta$) taken from different stages of the first epoch of CNN-A (mnist-back-image). Left: perturbations generated close to the beginning of training. Moving toward the right: perturbations generated closer to the end of the epoch. As the system evolves, $\delta$ mostly involves the portions of image covered by the digits.}
\label{fig:simpl-images-params}
\end{figure}

We qualitatively evaluated the  perturbation $\delta$ in Eq.~(\ref{eq:delta}), considering the CNN-A model of Table~\ref{tab:main-table}.
In Fig.~\ref{fig:simpl-images-params} we report the perturbations at different steps of the first epoch, applied to four randomly selected examples. The perturbation on the left concerns an example altered at the very early stages of the training procedures. The noise level decreases as long as learning proceeds (left-to-right in  Fig.~\ref{fig:simpl-images-params}), until we see that the system stops perturbing the background and some regions of digit area are emphasized. This is coherent with the fact that, as long as the network learns to focus on the digit area to discriminate the data, there is no need to alter the background anymore.

\begin{figure}

\begin{minipage}{0.5\textwidth}
        \begin{subfigure}[b]{0.48\textwidth}
          \centering
          \includegraphics[width=1.\linewidth]{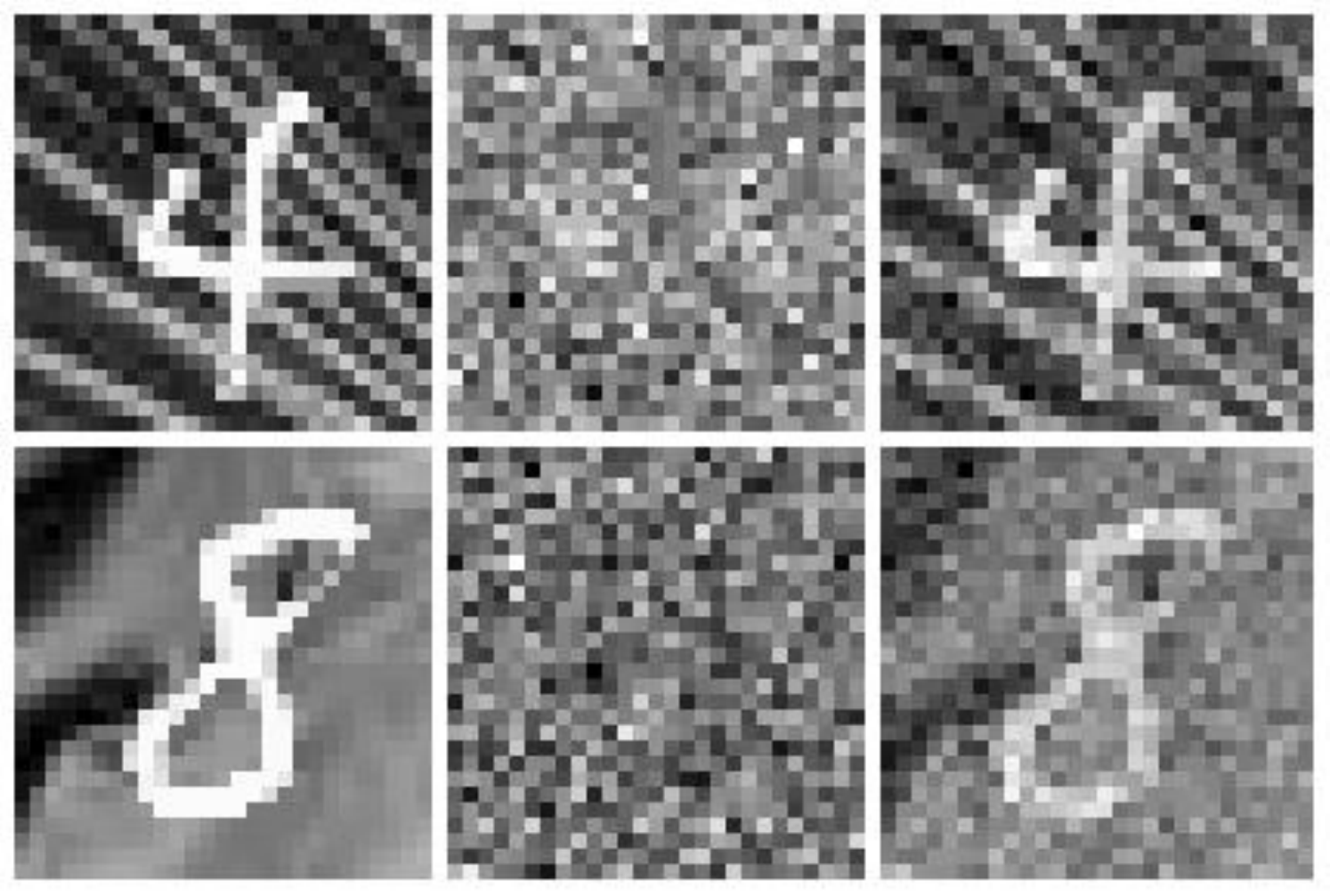}
          \caption{FC-A}
          \label{fig:ff-simpl-images}
        \end{subfigure}%
        \hfill%
        \begin{subfigure}[b]{0.48\textwidth}
          \centering
          \includegraphics[width=1.\linewidth]{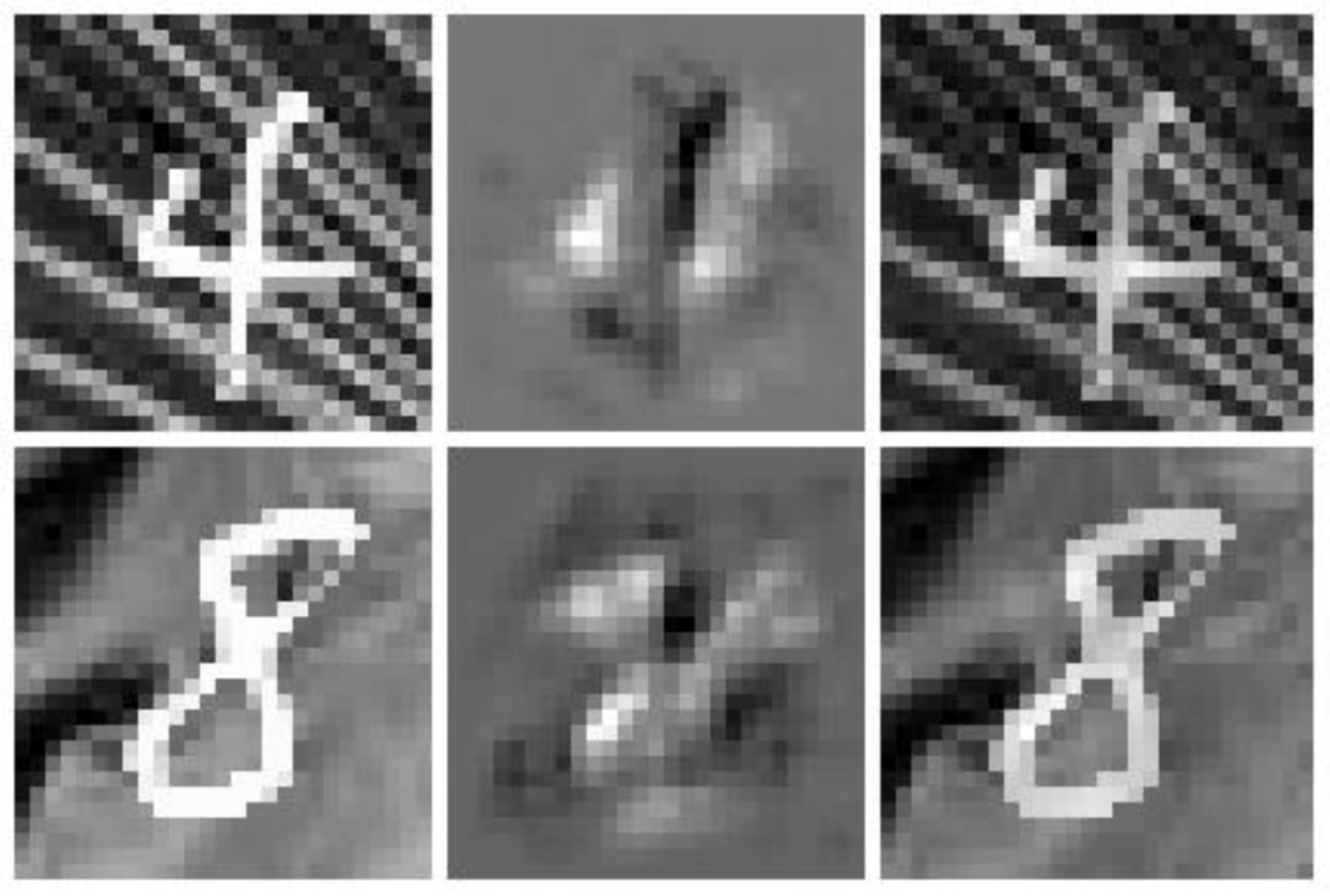}
          \caption{CNN-A}
          \label{fig:cnn-simpl-images}
        \end{subfigure}\\
        \vskip -2mm
        \begin{subfigure}[b]{0.48\textwidth}
          \centering
          \includegraphics[width=1.\linewidth]{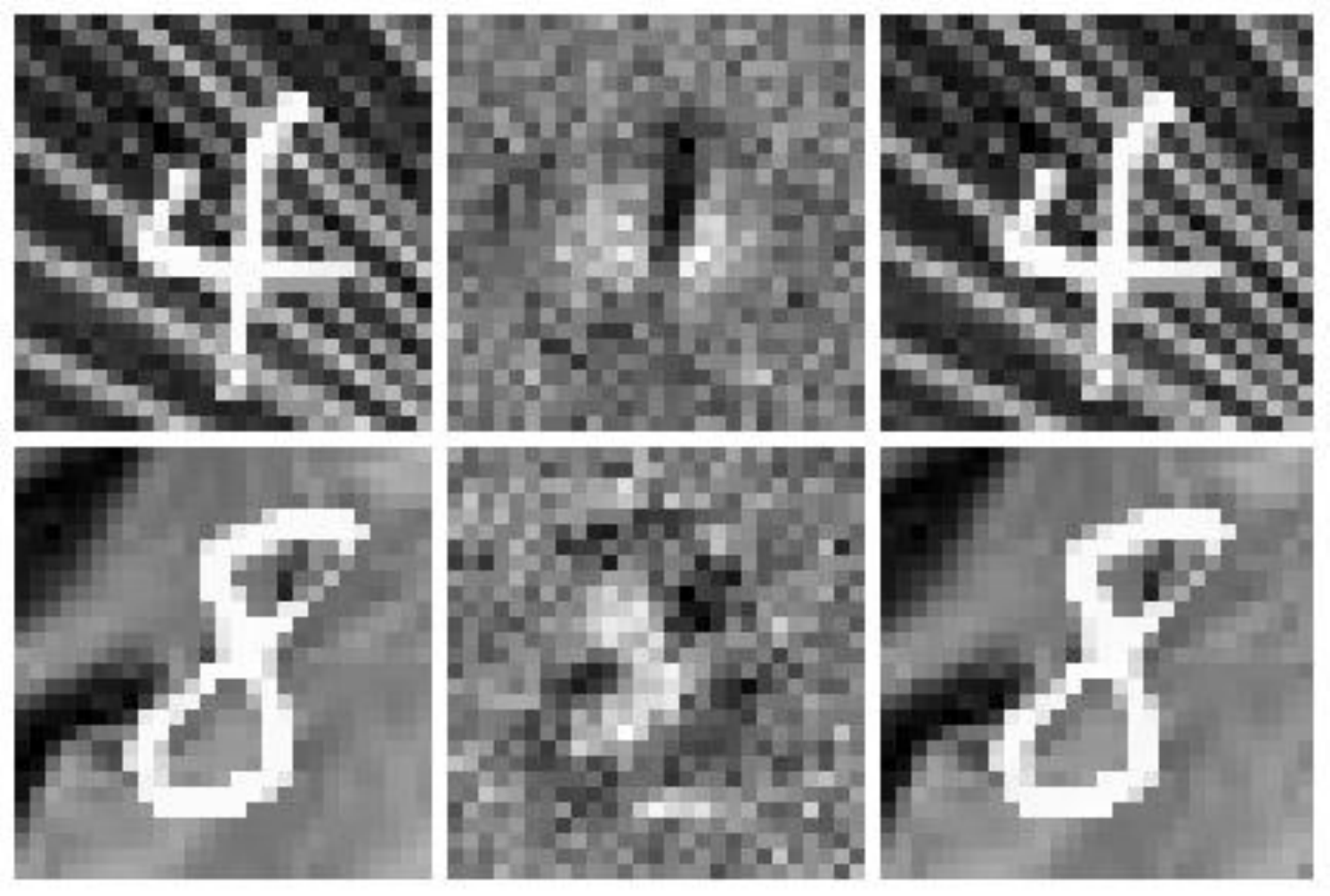}
          \caption{FC-B}
          \label{fig:ff2-simpl-images}
        \end{subfigure}%
        \hfill%
        \begin{subfigure}[b]{0.48\textwidth}
          \centering
          \includegraphics[width=1.\linewidth]{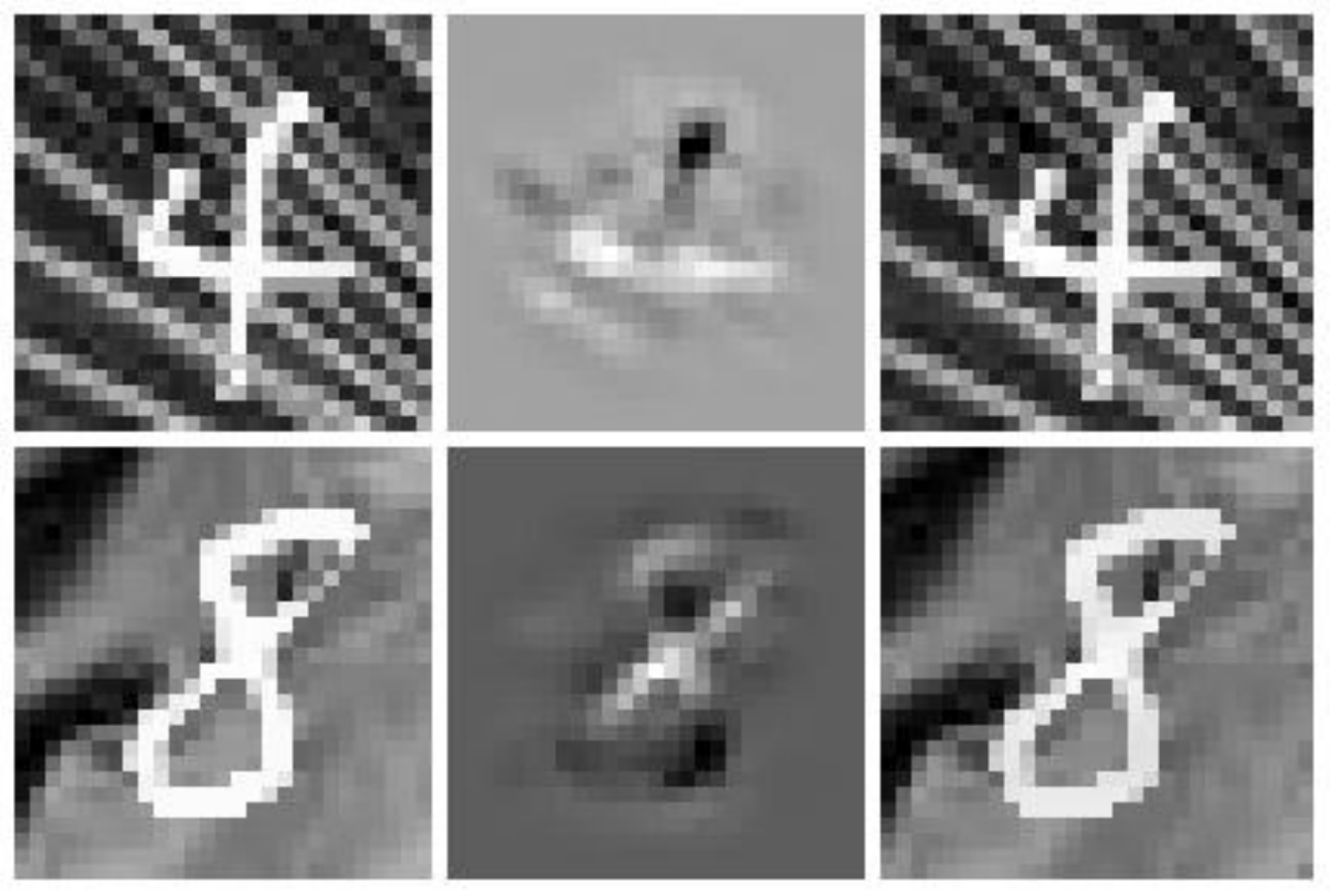}
          \caption{CNN-B}
          \label{fig:cnn2-simpl-images}
        \end{subfigure}%
\end{minipage}
\caption{Original data $x$ (first column), perturbation $\delta$ (second column - emphasized by normalization to make it more visible) and resulting ``simplified'' images $\tilde{x}$ (third column) for different networks (Table~\ref{tab:main-table}) at the end of the first epoch. Some simplifications are hardly distinguishable by a human.}
\label{fig:simpl-images}
\end{figure}

\begin{figure}[h]
\includegraphics[width=0.4\textwidth]{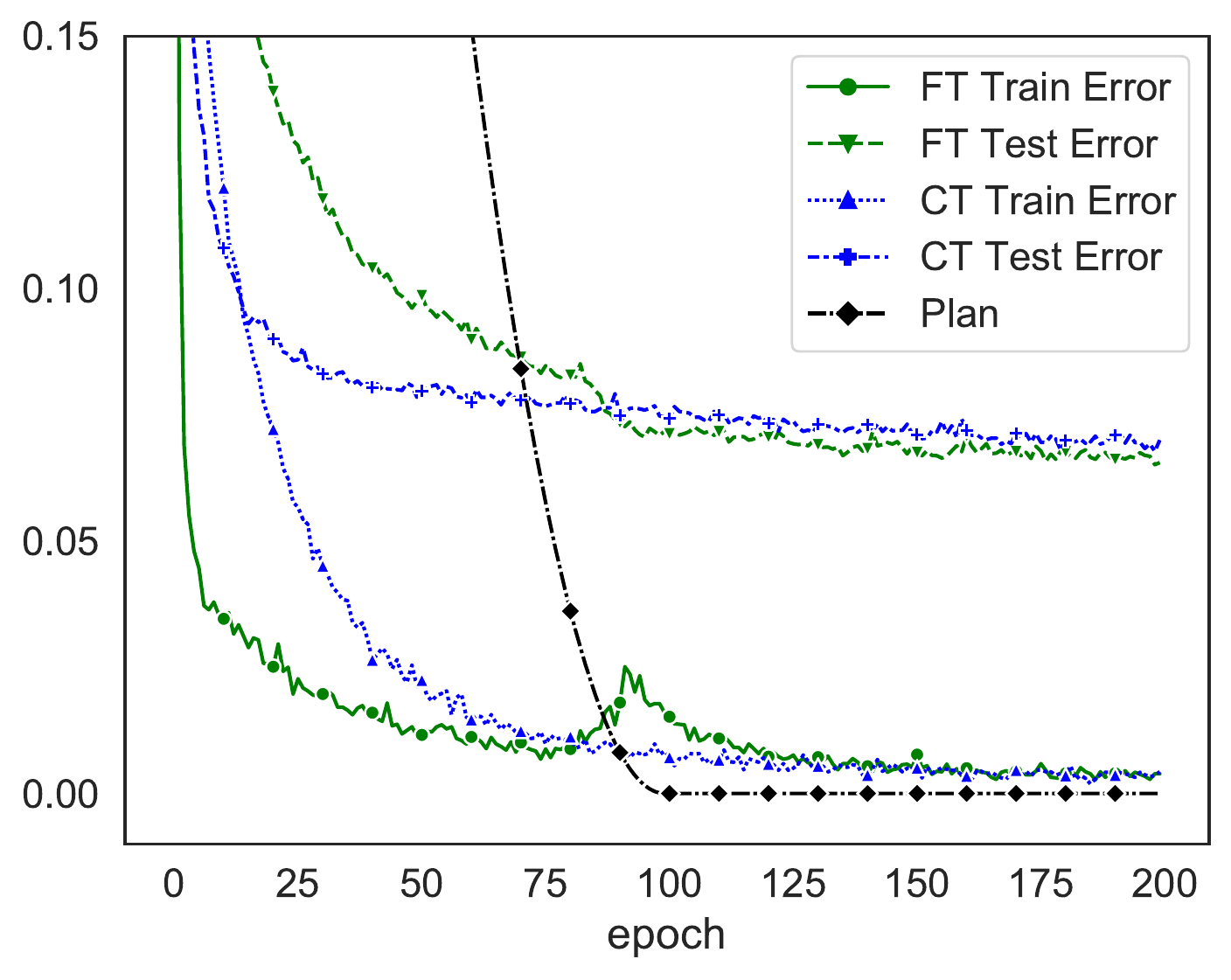}
\centering
\caption{Training and test error rates for FT and CT on a single run (mnist-back-image, CNN-A architecture). The black curve (plan) is proportional to $\tau^{\gamma}$.}
\label{graph}
\end{figure}

Then, we inspected the differences among the perturbations applied by the considered neural architectures, showing results in Fig. \ref{fig:simpl-images}. In order to compute the $\delta$'s, gradients are backpropagated by the FT algorithm across all the layers up to the input, and the network architecture plays an important role in shaping the perturbations. Interestingly enough, convolutional networks consistently provide more structured simplifications (Fig. \ref{fig:cnn-simpl-images}, \ref{fig:cnn2-simpl-images}), with slight manipulation of low-level local visual features in salient regions covered by the digits. On the other hand, $\delta$'s obtained from the FC-A and FC-B models (Fig. \ref{fig:ff-simpl-images}, \ref{fig:ff2-simpl-images}) are very noisy, with a limited emergence of visual structure in the deeper model FC-B (\ref{fig:ff2-simpl-images}). However, still most of the pixels get altered.
This rowdy perturbation might be the reason behind the limited success of FT in the fully-connected networks, and suggests directions for future improvements. As a matter of fact, the simplified examples get very artificial and far from the distribution of the original training data.
In Fig.~\ref{graph}, we report the evolution of the error rate during the training epochs (mnist-back-image, CNN-A), comparing FT and CT. We also report a black curve that is proportional to $\tau^{\gamma}$, thus showing how the developmental plan reduces the impact of the perturbation (when it becomes zero, data are not altered anymore).
 The small bump right before 100 epochs is due to the final transition from altered to original data. The test error of FT is higher than the one of CT when the data are altered, as expected, while it becomes lower when the simplification rate vanishes. 

\begin{figure}[h]
\includegraphics[width=0.24\textwidth]{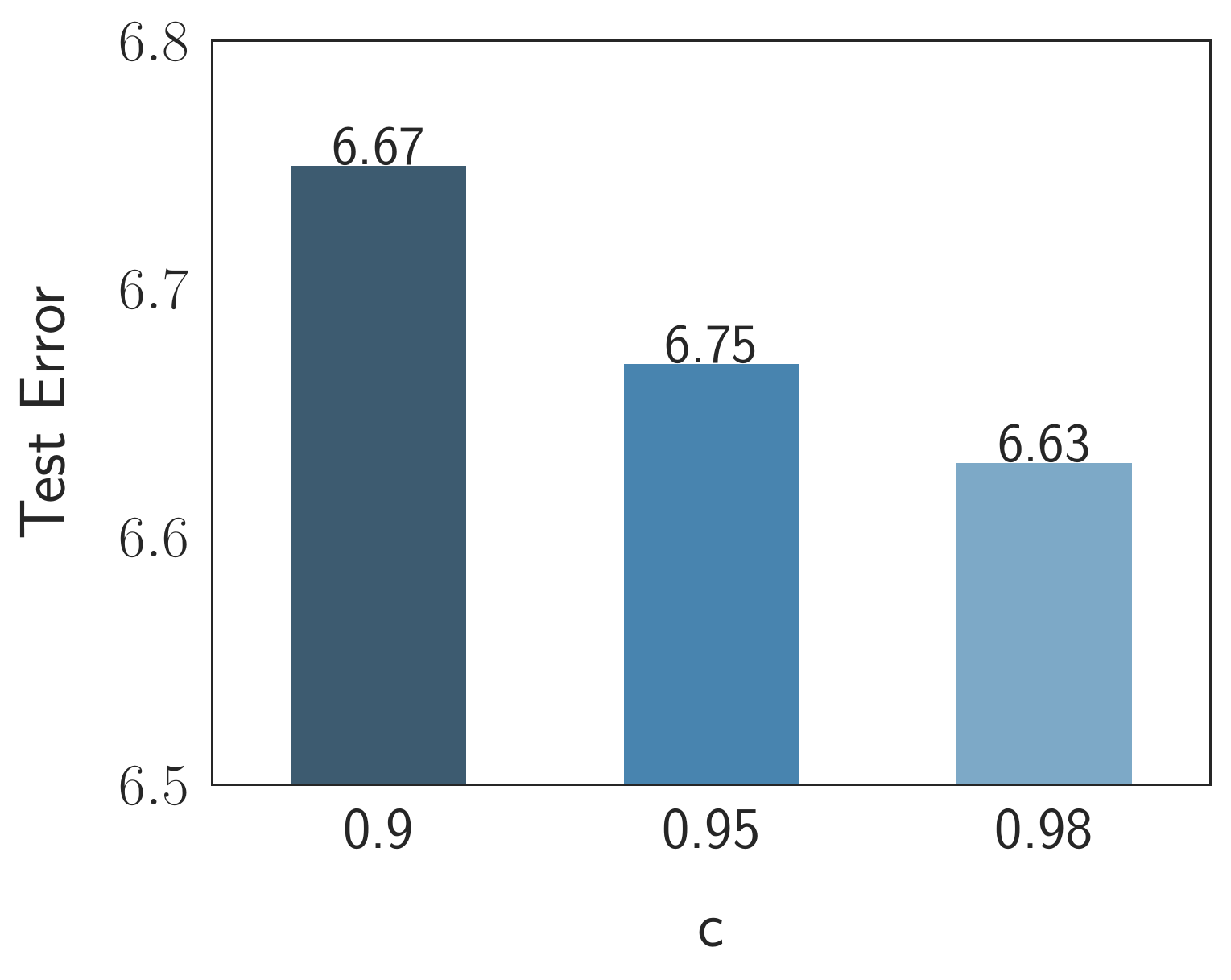}
\includegraphics[width=0.24\textwidth]{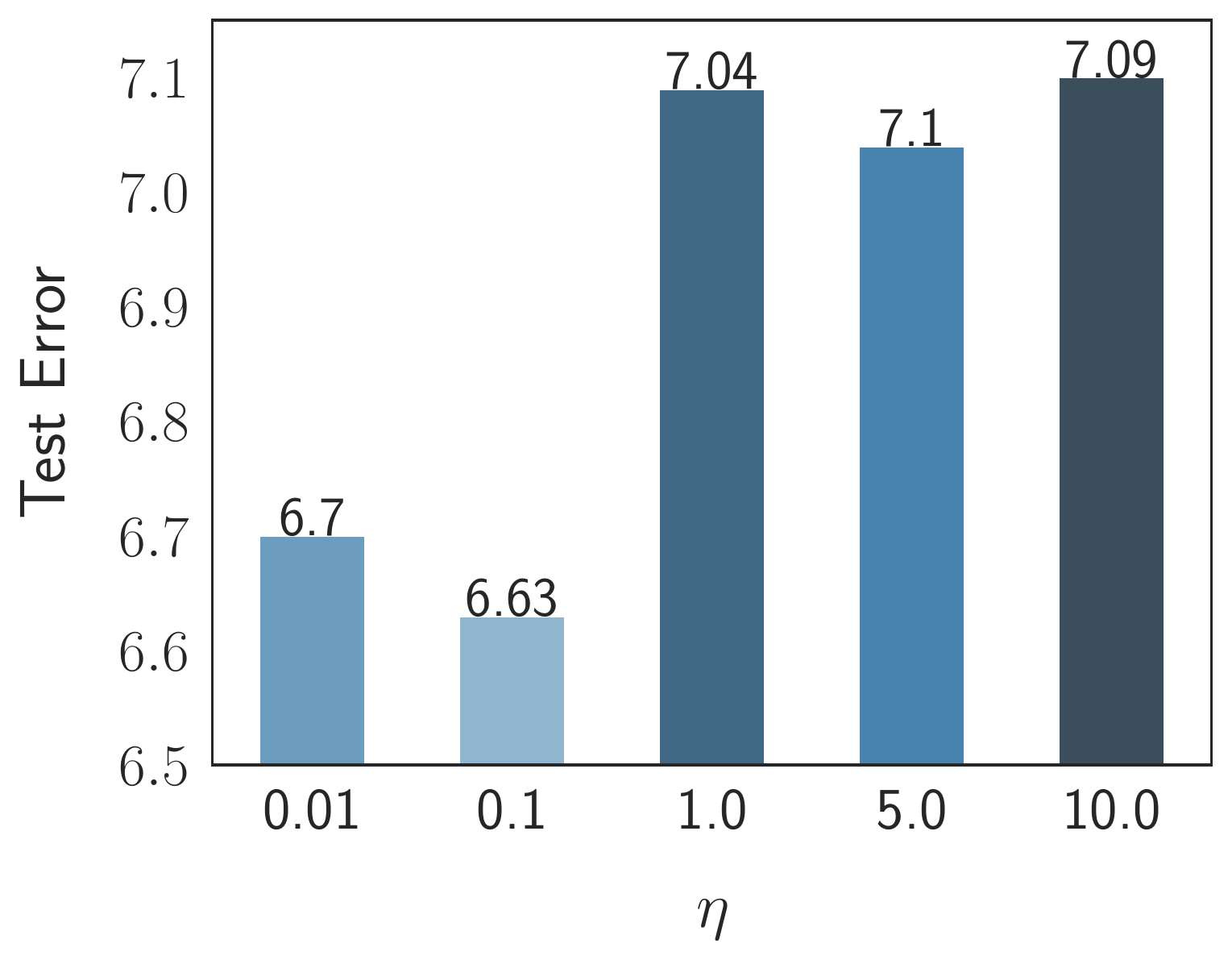}
\includegraphics[width=0.24\textwidth]{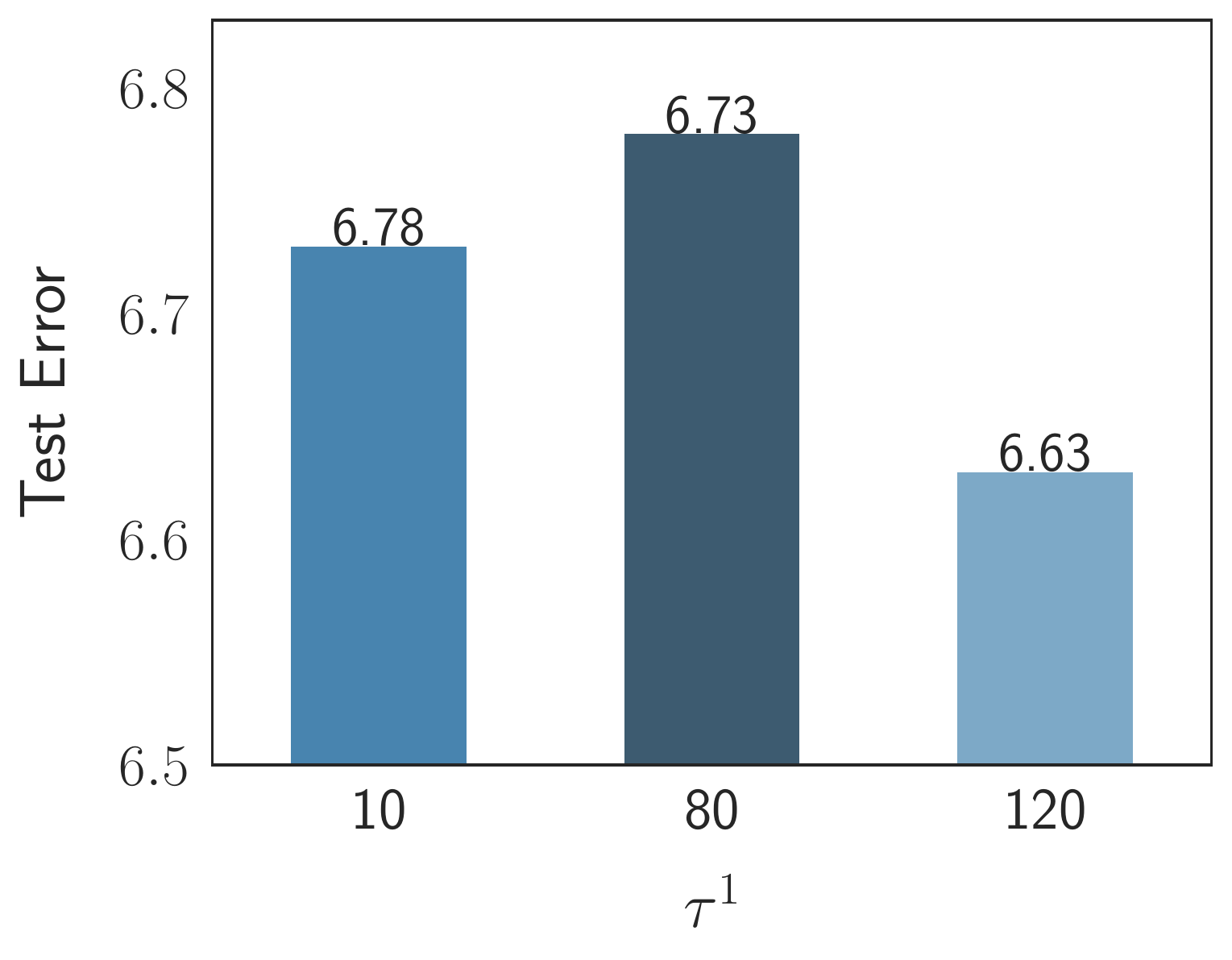}
\includegraphics[width=0.24\textwidth]{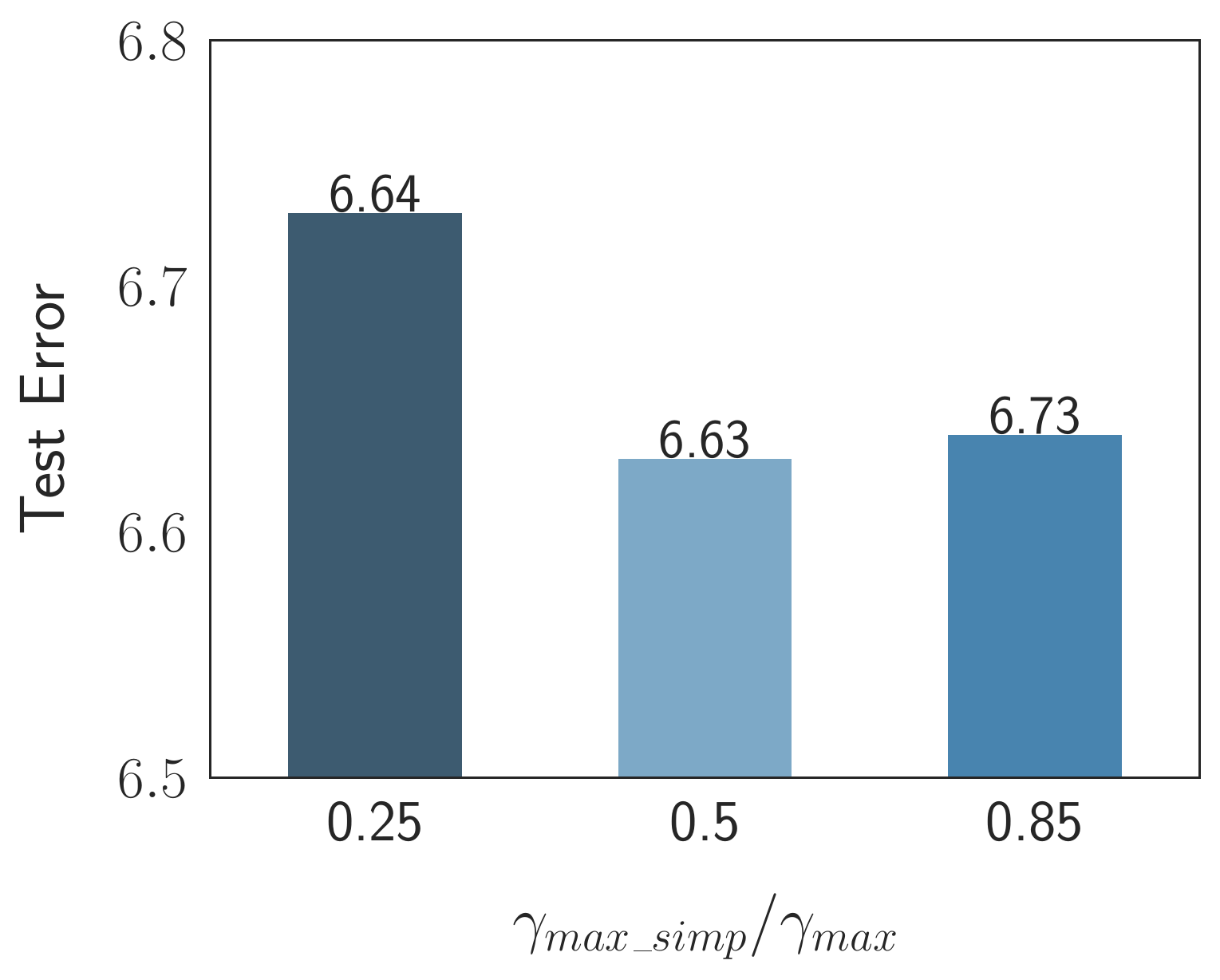}
\centering
\caption{Test error (single run) of CNN-A (mnist-back-image dataset) under different configuration of the FT hyper-parameters (see the paper text for a description).}
\label{histo}
\end{figure}

We also evaluated the sensitivity of the system to the main hyper-parameters of FT. In Fig.~\ref{histo}, we report the test error of CNN-A, mnist-back-image dataset, for different configurations of $c$, $\eta$, $\tau^1$ and $\frac{\gamma_{max\_{simp}}}{\gamma_{max}}$ in a sample run that is pretty representative of the general trend we observed in the experiments. Larger values of $c$ reduce the frequency of early-stops in computing the perturbations, thus allowing FT to alter the data more extensively and improve the performance. The preference on the largest value of $\tau^{1}$ suggests that an aggressive perturbation at the early stages of learning helps. Small learning rates ($\eta$) allow the system to avoid extreme alterations of the data, while a developmental plan covering at least $50\%$ of the training steps seems more effective than the shorter-term plans, $\frac{\gamma_{max\_{simp}}}{\gamma_{max}} = 0.5$.

\section{Conclusion and Future Work}
\label{sec:conclusions}
We presented a novel training procedure named Friendly Training, that allows a network to alter the training data accordingly to a developmental plan, in order to implicitly learn from more manageable data. Differently from related work, the network decides what information to discard from the data at different stages of the training procedure, leading to improved generalization skills and a smoother development of the decision boundaries. We plan to extend this approach introducing a separate neural model to estimate the simplification that should be applied to the data, instead of directly optimizing the perturbations.

\bibliography{biblio}
\bibliographystyle{IEEEtran.bst}


\end{document}